# Post-processing Multi-Model Medium-Term Precipitation Forecasts Using Convolutional Neural Networks

Authored by Bob de Ruiter
Supervised by Yuliya Shapovalova
Advised by Kirien Whan (KNMI) and Maurice Schmeits (KNMI)

**Abstract**

The goal of this study was to improve the post-processing of precipitation forecasts using convolutional neural networks (CNNs). Instead of post-processing forecasts on a per-pixel basis, as is usually done when employing machine learning in meteorological post-processing, input forecast images were combined and transformed into probabilistic output forecast images using fully convolutional neural networks. CNNs did not outperform regularized logistic regression. Additionally, an ablation analysis was performed. Combining input forecasts from a global low-resolution weather model and a regional high-resolution weather model improved performance over either one.

# Contents

# Acknowledgements

Thanks to Yuliya Shapovalova for the encouragement and patience, as well as for the thorough feedback on my thesis every step of the way. Thanks to Kiri Whan and Maurice Schmeits of KNMI for contributing their expertise in frequent video meetings, and for their detailed comments on the drafts.

Thanks to Iwan Holleman for an informative meeting and for introducing me to Maurice Schmeits.

Thanks to my friends and family for the unwavering support.

# 1 Introduction

Reliable precipitation forecasts are important to businesses, governments, and private citizens alike. Precipitation forecasts are used to inform decisions on everything from agriculture to road safety (e.g. Mjelde et al. 1988).

Numerical weather prediction is typically performed by estimating the initial conditions, simulating future weather developments using a deterministic atmospheric model, and post-preprocessing the simulation using statistical methods. Since most operational atmospheric models are deterministic (e.g. Bengtsson et al. 2017), it can be difficult to quantify the degree of uncertainty of the forecasts. Because of this, in addition running a single deterministic forecast, most national weather services also run an ensemble forecast, which itself consists of a large number of deterministic forecasts using slightly distorted but realistic initial conditions. These perturbations can have quite a large effect on the simulated outcome. As such, the distorted forecasts can be used to get a sense of forecast certainty. In forecasts produced for the general public, probability of precipitation (POP) at a given location are preferably predicted using an ensemble model, although it is possible to post-process deterministic forecasts (Theis, Hense, and Damrath 2005). The simplest way to obtain a probabilistic forecast from an ensemble, known as raw ensemble forecasting, is to compute the percentage of ensemble members forecasting precipitation above a predefined threshold. For example, if two out of ten ensemble members forecast above-threshold precipitation, the reported POP would be 20%.

Ensemble members, however, are biased towards being similar to an undistorted forecast. As such, ensembles do not capture all uncertainty, and raw ensemble forecasts are generally overconfident (Wilks 2019). To correct biases and dispersion errors in NWP models, statistical post-processing is employed. For instance, forecasts can be calibrated using isotonic regression (Barlow and Brunk 1972). Alternatively, machine learning models can be trained to combine and correct forecasts (Wilks 2019).

Members of an ensemble forecast are usually composed of runs from the same NWP model, only differing in initial conditions. However, there is nothing preventing the combination of forecasts from different NWP models. Multi-model ensembles (MMEs) have been shown to outperform single-model ensembles on multiple meteorological variables, among which 24-hour precipitation accumulations (Mylne, Evans, and Clark 2002).

In contrast to homogenous single-model ensembles, where usually no a priori reason exists to trust one of the ensemble members over the other, the track record of the members of an MME can be used to determine their accuracy and, therefore, their optimal weight in the multi-model ensemble. This can be achieved by, for instance, fitting a linear regression model on historical forecasts and observations, or logistic regression in the case of probabilistic forecasts.

Using the historical precipitation forecasts of eight operational models of government-funded weather services as training data, Krasnopolsky and Lin (2012) found that neural networks outperform linear regression when time and location data were added to the feature set. The authors emphasize the non-linearity of the model when explaining the increase in performance relative to linear regression. Although not discussed by the authors, the fact that non-linearity allows the neural network to learn location-specific weights could be a contributing factor to performance.

Multi-model ensembles for precipitation typically take into account the predicted precipitation in the neighborhood. This would be desirable: for example, if one model in the ensemble predicts that in 24 hours from now, a rainstorm will closely miss a grid cell, and historically, whenever that model made a similar prediction, the rainstorm did hit the valley, then the MME could correct for this bias. The same holds for biases with a temporal element: if one model in the ensemble consistently triggers convection too early, the combining model might be able to correct for this. Convolutional neural networks (CNNs) hold great potential to both combine member predictions into one forecast while correcting for spatial and short-term temporal biases. While CNNs have not yet been used for post-processing atmospheric models for precipitation, 2D CNNs and 3D CNNs have been used successfully as an end-to-end system for short-term predictions, outperforming state-of-the-art models based on optical flow such as ROVER (Shi et al. 2017). The same holds for recurrent CNNs, which outperformed both fully connected RNNs and ROVER (Shi et al. 2015). End-to-end systems only use a short-term history of precipitation observations as an input, rather than input forecasts from atmospheric models.

There appears to be no peer-reviewed work comparing CNNs used as end-to-end-systems directly to state-of-the-art atmospheric models for probabilistic forecasts of medium-term precipitation. Agrawal et al. (2019) does find that one atmospheric model, HRRR, outperforms one particular end-to-end CNN architecture, U-net, when the forecast window extends beyond five hours. Both are deterministic forecasts for precipitation amount, however.

As CNNs are capable of modeling short-term precipitation developments on their own, a CNN-based post-processing model may be able to correct systemic errors that can't be corrected by simpler models that are unable to account for spatial and temporal biases. Additionally, CNNs might lend their own short-term predictive power to the ensemble of medium-term forecasts. This is especially plausible for the shorter lead times.

While it can be useful to model temporality explicitly in an end-to-end predictive system, such as including recurrent connections (c.f. Shi et al. 2015, 2017), a machine learning model post-processing forecasts from atmospheric models might not need this, as the time steps are already simulated in the input forecasts. 2D CNNs without a temporal element have already been used for the post-processing of medium-term wind forecasts in the Netherlands (Veldkamp 2020). Additionally, while 2D CNNs perform worse than recurrent CNN architectures as end-to-end systems for short-term precipitation forecasting, they do outperform models based on optical flow, which are commonly in use today (Shi et al. 2017).

In the field of data science, training a learning algorithm to combine the predictions of other learning algorithms is known as model stacking (Wolpert 1992). The combiners used in stacked ensembles might perform well as the post-processing component in multi-model ensembles, as they both combine highly correlated predictions into one output prediction. In both cases, logistic regression is considered a strong baseline (e.g. Krasnopolsky and Lin 2012).

In this thesis, CNNs are compared to random forests and logistic regression for combining and post-processing hourly precipitation forecasts 12 and 24 hours into the future. Output forecasts are probabilistic—the task is to forecast the probability of precipitation exceeding a threshold. In different subexperiments, this threshold is set to 0.5 mm/h, 1 mm/h, or 2 mm/h.

## 1.1 Feature Importance

Measuring the marginal benefit of each feature in post-processing helps identifying biases and limitations of input forecasts. It also helps relating the results to past work, and has the potential to inform future work.

Because input forecasts are highly collinear, traditional statistical measures of feature importance which assume independent predictors, such as interpreting the coefficients of logistic regression, are of limited use.

Permutation importance and random forest mean decrease in impurity (MDI) are gaining popularity in meteorological post-processing as a measure of the relative importance of features (e.g. in Eccel et al. 2007; Straaten, Whan, and Schmeits 2018). MDI is a feature importance metric calculated when fitting random forests, popular because it is cheap to compute and quite good at handling noisy collinear features. In this thesis, I argue that these metrics are of limited use when evaluating the usefulness of highly-correlated meteorological features. MDI is compared to the results of a series of ablation analyses. In an ablation analysis, an experiment is run twice, once with and once without a feature, and the results are compared. Since this is quite computationally expensive, ablation analyses are performed using logistic regression.

## 1.2  Meteorological Definitions

This thesis is read by both meteorologists and data scientists. Definitions for terms used by meteorologists are provided here for data scientists. Models used in data science are described in the *Methods* section. For a more extensive treatment of meteorological terms, see Wilks (2019).

The terms *input forecast* and *output forecast* are used in this thesis to differentiate between the forecasts used as the input of a post-processing model and the combined forecast output by a post-processing model. Some input forecasts are derived forecasts, such as the GEFS ensemble mean and the ensemble's first and third quartiles.

The *initialization time* is the moment from which the weather simulation runs forward. Observations reported up to the initialization time can be used in the simulation. The *verification time* of a forecast is the time the event being forecast has completed.

The *lead time* of a forecast is the difference between the initialization time and the verification time. For instance, the lead time of a forecast started at 00:00 UTC for the period 00:00 - 01:00 UTC has a verification time of 01:00 UTC and a lead time of +1h. In this thesis, the lead time in hours is commonly abbreviated as $l$.

A *reforecast*, *hindcast* or *retrospective forecast* is a forecast made after the event being forecast has occurred. The model is not given any more information than it would have had if making true forecasts. Reforecasts are used in meteorology to evaluate and calibrate models. Historical true forecasts are often less suitable for calibration because atmospheric models and their hyperparameters are changed throughout time in response to new insights, whereas reforecasts use a fixed atmospheric model and a single hyperparameter configuration.

A *run* is a set of forecasts made at the same initialization time with different lead times.

All forecasts and observations are provided as 2D grids. There are many ways to project spherical data onto a 2D surface. In this context, *reprojecting* means warping a map to match a new projection. *Regridding* is an operation consisting of interpolating one grid onto another grid.

A *control forecast* is often included in ensemble forecasts. This is a forecast with undistorted initial conditions and hyperparameters.

The terms *upscaling* and *downscaling* are not used in this thesis, since they have opposite meanings in meteorology and computer science. Instead, I speak of increasing and decreasing the resolution. *Reliability diagrams* and *calibration curves* are two terms for the same thing.

*Short-term forecasting*, also known as *nowcasting*, is defined in this thesis as producing forecasts with a lead time under 3 hours. Nowcasting models usually have a high temporal resolution. Short-term forecasting cannot take too long if the forecasts are to have any practical use.

*Medium-term forecasting* is defined in this thesis as producing forecasts with a lead time between 3 hours and 3 days. Operational medium-term forecasting systems use computationally heavy atmospheric models which, in many cases, take a couple of hours to produce a run of forecasts.

Finally, in the context of binary classification, the *precipitation cover* is the fraction of atomic observations which exceed the precipitation threshold.

## 1.3 Climatology of the Netherlands

The Netherlands has a temperate maritime climate. Almost all precipitation is rainfall.

Even though the Netherlands only has 41000 km$^2$ land mass, precipitation patterns are quite different across the country. In the coastal regions, the lion's share of precipitation falls in the winter. In the east, most precipitation falls in the summer (Figure 1).

Table 1: Summary of precipitation datasets

|  | 3h GEFS reforecast | HARMONIE reforecast | KNMI observations |
| --- | --- | --- | --- |
| Forecast frequency | Daily (00:00 UTC) | Daily (00:00 UTC) | - |
| Forecast window | 96 hours | 48 hours | - |
| Map extent | Global | West-Europe | Dutch land mass |
| Period | 1984-2020 | 2015-2017 | 2008-2020 |
| Resolution | ~25 km | ~2.5 km | ~1 km |
| Pixels on Dutch land | ~60 | ~6000 | 38028 |
| Variable used | Precipitation | Rain | Precipitation |
| Time step | 3 hours | 1 hour | 1-hour accumulations |
| Accumulation window | 6 hours | 48 hours | 1 hour |
| Projection | lat/lon | Rotated lat/lon | Polar stereographic |
| Ensemble (members) | Yes (10 + control) | No | - |

# 2 Data

## 2.1 Collection

Input forecasts were collected from 2 governmental reforecasting projects operated by governmental weather services.

Hourly HARMONIE rain reforecasts with a 48-hour forecast window were kindly provided by KNMI, the Dutch national weather service, for use in this thesis. HARMONIE, the most recent model by the international consortium HIRLAM, is used for the KNMI's medium-term weather forecasts. The reforecasts were originally produced for an unrelated project. As such, the HARMONIE reforecasts do not aim to forecast non-rain precipitation, such as snow and hail, and are limited to the 2015-2017 period. The model is described in detail by Bengtsson et al.

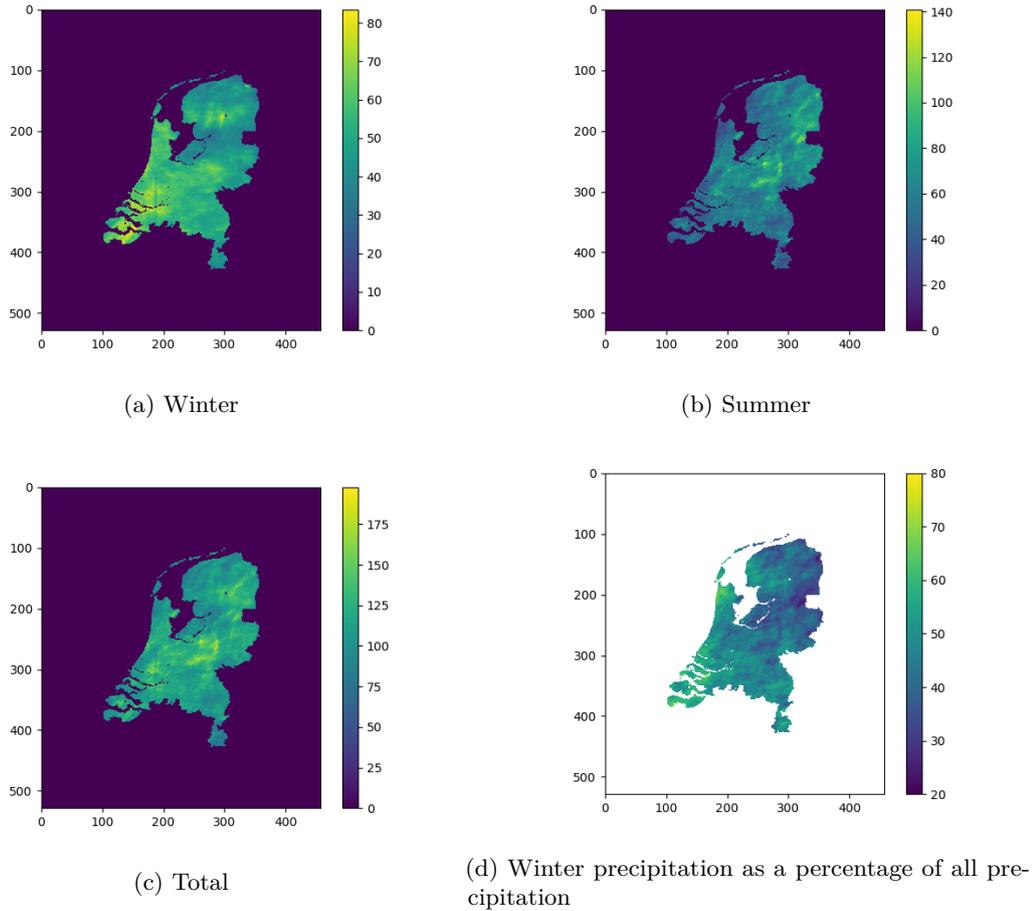

Figure 1: Seasonality of precipitation differs across the Netherlands (12 UTC, corresponds to $l = 12$ forecasts). Data from 2015 to 2017. Note that color coding is not consistent across subfigures.

(2017). The 40th cycle of the model is used.

Three-hourly precipitation reforecasts, part of the output from the second-generation Global Ensemble Forecast System (GEFS), were obtained from the Earth System Research Laboratory, part of the American National Oceanic and Atmospheric Administration. The temporal extent of the GEFS reforecast spans from 1984 to, at the time of writing, just a day ago. This data, described in detail by Hamill et al. (2013), is publicly available[1].

The collection of precipitation observations deemed most reliable by the KNMI, the hourly accumulative radar dataset calibrated and corrected by rain gauges, were collected from the KNMI open data repository. The KNMI accumulations are limited to Dutch land mass. The ground truth used in this thesis is imperfect; radar images are known to be noisy. Station observations are known to be more accurate, but the spatial coverage of radar is much better. One radar image has 38028 valid atomic observations for Dutch land mass. By contrast, KNMI operates less than 40 weather stations, although there are also 300 weather stations run by volunteers.

---
[1] https://psl.noaa.gov/forecasts/reforecast2/download.html

High buildings interfere with radar observations at select locations in the Netherlands (Straaten, Whan, and Schmeits 2018). Observations for these locations are not used in model training and evaluation.

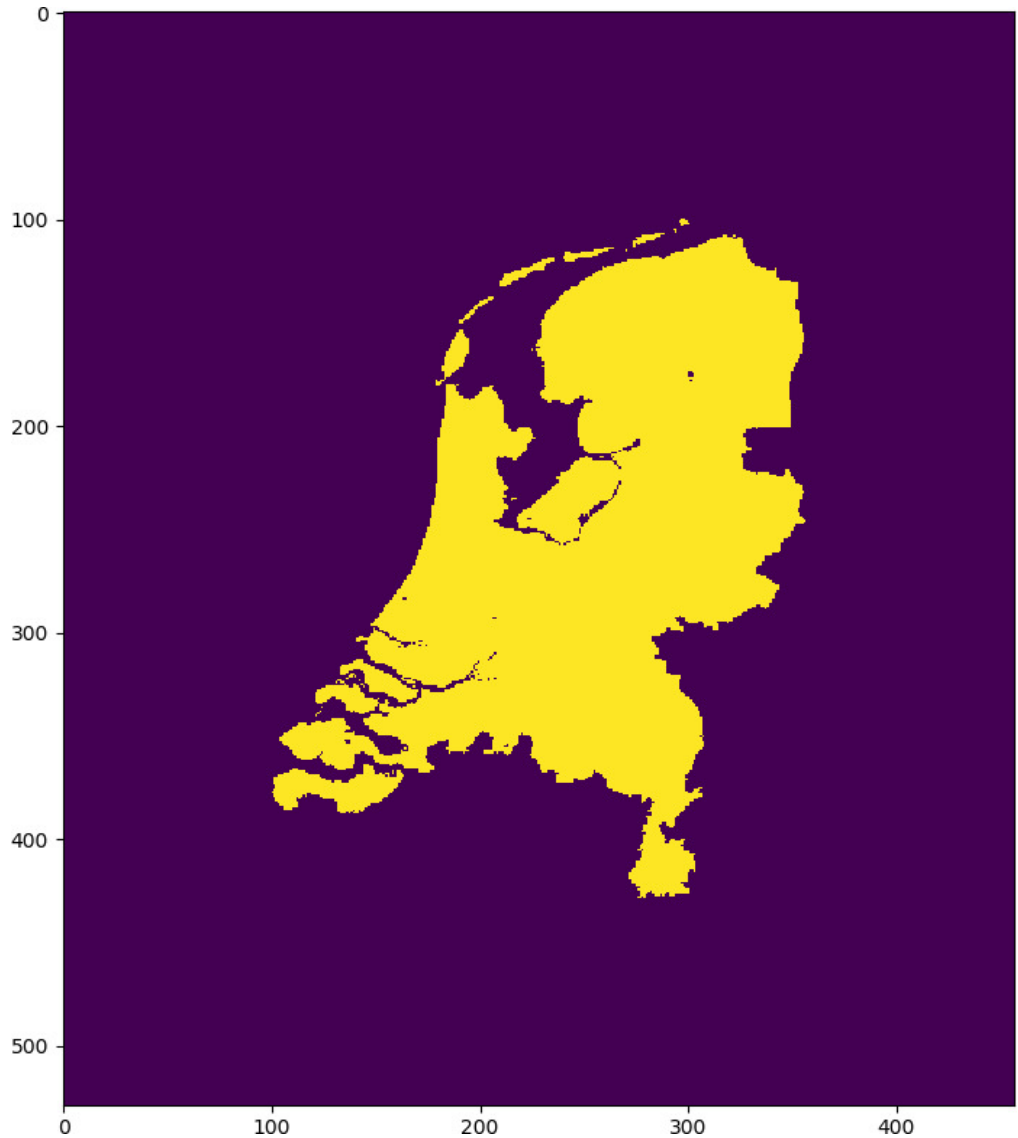

Figure 2: Locations for which valid observations were provided. Holes in the mask correspond to tall structures interfering with weather radars.

## 2.2 General Preprocessing

Four input forecasts were obtained from the GEFS reforecast by taking the ensemble mean, the ensemble first quartile, the ensemble third quartile, and the GEFS control forecast. Here, the control forecast is also considered an ensemble member, bringing the total number of ensemble members to 11. This means that the first quartile is the third-lowest prediction and the third quartile is the third-highest. Adding the HARMONIE forecast, this results in five sets of input forecasts which each have five dimensions: forecast validation time, lead time, a spatial x-coordinate, and a spatial y-coordinate. All member forecasts were cropped and reprojected to the observation grid, which uses a polar stereographic projection.

Because all input forecasts have a lower spatial resolution than the 1 km resolution of the observations, the forecast resolution was increased using nearest neighbor interpolation. I hypothesized that CNNs have a comparative advantage when forecasting high-resolution images using lower-resolution forecasts, as CNNs can learn to perform superresolution (Kim, Kwon Lee, and Mu Lee 2016).

The remainder of the preprocessing, training and testing pipeline is applied to each lead time separately, unless stated otherwise. As such, each set of input forecasts, referred to as $G_m$ (the GEFS ensemble mean), $G_1$ (the first quartile), $G_3$ (GEFS third quartile), and H (the deterministic HARMONIE forecast), is considered to only have three dimensions from now on: validation time $t$, x-coordinate $i$, and y-coordinate $j$. Like every input forecast $F$, the observed precipitation $O$ has three dimensions: observation time $t$, x-coordinate $i$, and y-coordinate $j$, such that $F_{t,i,j}$ is a forecast for $O_{t,i,j}$. The temporal dimension is limited by the availability of HARMONIE data, which runs from 2015 to 2017 and consists of 1096 daily reforecasts.

In addition to the HARMONIE forecast corresponding to the lead time of the output forecast $l$, forecasts for the two hours before and after were included as input forecasts. Similarly, for GEFS, which has a time step of 3 hours, the ensemble means for the three-hour blocks before and after were used as input forecasts.

A local maximum forecast was derived from the GEFS ensemble mean: for each pixel, the highest forecasted precipitation in a 100x100km square window centered on that pixel was computed. The extent of the forecasts was much higher than the extent of the observations, so no zero-padding was required at any of the borders. The local maximum of precipitation forecast was thought to be especially relevant for longer lead times, as translation errors increase in both size and prevalence as the lead time increases.

One feature, the GEFS raw ensemble forecast, depended on the specifics of the classification task. It was computed by taking the fraction of GEFS ensemble members that forecast a precipitation amount higher than the ground truth threshold. For instance, when the task consists of predicting whether precipitation exceeds 2 mm, the raw ensemble forecast is the fraction of GEFS ensemble members which forecast 2 mm or more.

Non-forecasts predictors were also computed, all of the same shape as $O_{t,i,j}$. This includes the observed precipitation in the hour leading up to the forecast initialization; the error of the previous forecast of the same lead time for both HARMONIE and the GEFS ensemble mean; the x-coordinate and the y-coordinate on the observation grid; and the cosine of the day of year. When making temperature forecasts for Germany using feedforward neural networks, including longitude, latitude, and altitude has also been found to improve performance, although this seemed to mostly be the result of including altitude (Rasp and Lerch 2018). The Netherlands are much flatter than Germany. As such, altitude was not included as a feature.

The day of year was transformed such that January 1 and December 31 mapped to 1 and June 31 mapped to -1 (Equation 1).

$$tdim = cos\left(\frac{d * 2\pi}{365}\right) \quad (1)$$

The non-forecast predictors are likely only weakly correlated to precipitation, but there might be interactions with input forecasts. For example, one input forecast might perform better in coastal regions in the west while the other performs better in the east.

All features were standardized. A summary of all features, both input forecasts and other predictors, can be found in Table 2.

Since 2016 is a leap year, a total of 1096 observation images are used per lead time. However, some observed hourly accumulations required to train and evaluate the models were missing from the KNMI dataset, the files only containing undefined values. For the 12-hour and 24-hour lead time forecasts, respectively 1 and 2 radar images were missing. Corresponding forecasts were not used in training and evaluation.

Finally, a binary observation mask was defined. This mask was of the same shape and size as the observation grid. All grid cells which contained large buildings known to interfere with radar were masked out (Straaten, Whan, and Schmeits 2018), as well as grid cells not on Dutch land mass. A mask with 38028 cells remained (Figure 2).

Table 2: All features

| Feature | Description |
| --- | --- |
| harmonie | HARMONIE forecast for lead time $l$ |
| hm2 | HARMONIE forecast for lead time $l-2$ |
| hm1 | HARMONIE forecast for lead time $l-1$ |
| hp1 | HARMONIE forecast for lead time $l+1$ |
| hp2 | HARMONIE forecast for lead time $l+2$ |
| gefs_avg | GEFS ensemble mean |
| ga_prev | The GEFS ensemble mean in the forecast for the previous 3 hours* |
| ga_next | The GEFS ensemble mean in the forecast for the next 3 hours* |
| gefs_control | The GEFS control |
| gefs_q1 | GEFS first quartile |
| gefs_q3 | GEFS third quartile |
| gefs_t | the GEFS raw ensemble forecast |
| init_obs | the observed precipitation in the hour before initialization time |
| ydim | the y coordinate on the observation grid |
| xdim | the x coordinate on the observation grid |
| tdim | cosine of the day of year |
| harmonie_past_error | the error of the previous validated HARMONIE forecast of the same lead time |
| gefs_avg_past_error | the error of the previous validated GEFS ensemble mean of the same lead time |
| gefs_avg_lmax | GEFS ensemble mean local maximum in a 100x100 km window |

# 3 Methods

Probabilistic precipitation forecasting was modeled as a binary classification problem by setting a threshold, labeling all observations above that threshold as positive samples, and the remainder as negative. By training multiple instances of binary classifiers using different thresholds, a cumulative density function of precipitation amount could be estimated. The downside of this approach is that the number of models that have to be fitted scales linearly with the resolution of the cumulative density function. Because computational resources were limited, three precipitation thresholds $h$ were selected: 0.5 mm, 1 mm and 2 mm. For the same reason, all models were only trained and tested on 12-hour and 24-hour lead times.

Binary labels were obtained by applying the precipitation threshold (Equation 2).

$$Y = O > h \qquad (2)$$

0.5 mm was chosen as the lowest threshold because radars are known to be noisy. A lower threshold would cause some samples without any precipitation at all to be labeled as positives. It is almost certain that some samples near the precipitation threshold were mislabeled because of the noise in the observations. 1 and 2 mm thresholds were chosen because probabilistic forecasting for high thresholds is considered to be difficult for atmospheric models, perhaps leaving more room for improvement through post-processing. Forecasting heavy rain is relevant in many applications of weather forecasts.

Deciding on how to split meteorological data into training, validation and test sets is not trivial, since consecutive observations and forecasts are correlated, meaning that some information could leak back in time if cross-validation is used. In many forecasting pipelines outside the field of meteorology, the training set is limited to events that occurred before the test data. This would severely limit train and validation set size for the 2015 data, and it would make it near-impossible to use the first few months of 2015 as a test set. It is also overly cautious, since the information leak is limited. Precipitation in consecutive years is hardly correlated, and while it might be useful to know the weather in a week from now when forecasting the weather tomorrow, information about the weather two weeks from now provides very little information about the upcoming days.

For three-year medium-term forecasting, a commonly-used scheme for test set selection exists: three-fold cross-validation where one year of consecutive observations is used as a test set for each fold. This does not eliminate leakage entirely: the observations and forecasts of the first week of January 2017, for instance, will correlate to the observations and forecasts of December 2016. If a model tested on the 2016 data were to overfit on a pattern in the January 2017 data, it might do very well on the last week of 2016 despite generalizing poorly. In practice, this issue is minimal, since the block of test data is so large.

Using one year of data for testing, this leaves two one-year blocks for training and validation. While the selection of a test set should use a conservative, well-tested scheme as it directly affects the integrity and interpretability of the test results, the train-validation split allows for slightly more freedom. The validation set is used for early stopping, model selection, and model calibration. Although the exact train-validation split can greatly influence performance, it does not bias the final results in any way.

Two parameters of this split are important: the fraction of data used for validation and whether all validation data should be consecutive. As far as I know, these hyperparameters cannot be optimized fairly in any conventional way such as grid search, since they pertain to the validation set itself.

I opted to use every second non-test sample as validation data (Figure 3).

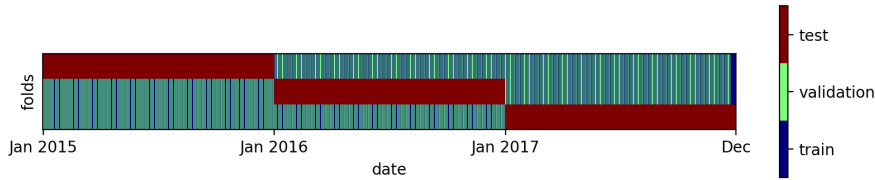

Figure 3: Train-validation-test splits for every fold

The other option, using one continuous block of validation data, might seem like a sensible choice for the same reasons it is a sensible choice for test data: very little information leaks from the training set to the validation set. This might ensure that the best-generalizing model is chosen, but this is not necessarily the case. One year of continuous validation data would likely be less diverse in terms of weather types and weather patterns than observations sampled from two years, as consecutive images are relatively likely to be of the same weather type. So while this validation scheme might promote overfitting on a sample level, it might also promote generalization across weather types. Not removing one continuous block from the non-test data increases the diversity of the training data as well, which might increase generalization as well.

One could take this idea even further. The baseline methods allow for placing part of an image in the training set and another in the validation set. I refrained from doing this, since it would further promote overfitting and make the comparison between CNNs and the baseline methods harder.

The choice to use half of all non-test data for validation was motivated by looking at the variance of possible evenly-spaced validation sets, as the decision to use every $n$th image for the validation set prevents stratification based on labels. The $h = 0.5$ mm, $l = +12$ test set was used to verify this decision, as was the $h = 2$ mm, $l = +12$ test set. Especially for the 2 mm threshold, there was no way to calibrate the forecasts using a smaller fraction of data.

Publications by the KNMI differentiate between winter and summer precipitation (e.g. Beckmann and Adri Buishand 2002; Straaten, Whan, and Schmeits 2018). Summer is defined as spanning the period from April 15 to October 14. Winter, then, runs from October 15 to April 14. In the Netherlands, winter precipitation patterns are quite different from summer. For instance, during summer, short bursts of heavy rain are relatively common.

During summer and winter, the error characteristics of NWP models also differ. As such, besides training each post-processing model on both summer and winter, I follow Straaten, Whan, and Schmeits (2018) in training and testing separate models on just winter and just summer. Experiments are run separately for each combination of lead time (+12, +24), threshold (0.5, 1, 2 mm), and season (winter, summer, all), unless stated otherwise, bringing the total number to 18.

Table 3: Percentage of valid atomic observations labeled as positive, i.e. where the precipitation exceeds the threshold (2015-2017, whole year)

| Lead time (h) | Threshold (mm) | Climatological probability |
|---|---|---|
| +12 | .5 | 5.5% |
|  | 1 | 3.0% |
|  | 2 | 1.2% |
| +24 | .5 | 5.0% |
|  | 1 | 2.8% |
|  | 2 | 1.1% |

All output forecasts were calibrated on the validation set using isotonic regression (Barlow and Brunk 1972). Any prediction outside the range of predictions for the validation set was clipped to the minimum or maximum prediction for the validation set. The reported validation scores are for the uncalibrated model. The reported test scores are all calibrated.

## 3.1 Pixel-by-pixel forecasts

In the baseline method, output forecasts for an atomic thresholded observation $Y_{tij}$ (i.e. for one particular location and time) only depend on $F_{tij}$ for each input forecast $F$. Note that one input forecast does provide information about neighboring pixels. Neighborhood information is provided through the GEFS local maximum computed in the preprocessing step. A few input forecasts provide information for other verification times. Forecasts for the same pixel at adjacent lead times are named `hm2`, `hm1`, `hp2`, `hp1`, `ga_prev`, and `ga_next` (see Table 2 for descriptions).

As the baseline methods, unlike CNNs, do not depend on the spatial layout of the atomic observations and features beyond post-processing, the feature set was transformed into tabular data, with an atomic observation in each row and a feature in each column. Since the reference observations are limited to Dutch land mass and observations for grid cells containing tall buildings are unreliable, forecasts and observations not in the observation mask (Figure 2) were discarded.

To be more detailed, each three-dimensional feature $F$, as well as the label matrix $Y$, was flattened along all dimensions: time $t$, x-coordinate $i$, and y-coordinate $j$. The observation mask was applied to both the features and the labels. This resulted in 19 forecast vectors and a label vector $y$. Since all datasets are flattened in the same fashion, $f_d$ remains a forecast for $y_d$, where $f$ is the feature vector obtained by flattening $F$ and $y$ is the label vector obtained by flattening $Y$.

The flat forecast vectors were combined into a feature matrix $M$, which is used by all baseline pipelines.

### 3.1.1 Logistic Regression

Logistic regression is a linear model closely related to linear regression. The intercept scalar $\alpha$ and the coefficient vector $\beta$ are fit according to Equation 3.

$$logit(o) = log(\frac{o}{1-o}) = \alpha + \beta \cdot M \qquad (3)$$

Here, the log loss is minimized (Equation 4).

$$logP(yt|yp) = y_t log(y_p) + (1 - y_t)log(1 - y_p) \qquad (4)$$

The logit transformation ensures *o*, which lies between 0 and 1, can be interpreted as the probability of positive class membership if the model is fit using 0 and 1 as target values for the negative and positive class, respectively. Probabilities resulting from logistic regression are well-calibrated if the logit of the probability is linearly related to the features. In practice, predictions often remain relatively well-calibrated even if the assumption of linearity is violated.

High collinearity (i.e. high correlation between features) can lead to unreliable and unstable coefficients, making the coefficient values less useful as a measure of feature importance. In the paradigm of machine learning, where logistic regression is used as a predictive model more than as a descriptive model, collinearity is less of an issue for logistic regression. Logistic regression performs well as an ensemble model, even though the predictions of base models are often highly correlated. For example, the winner of the 2015 edition of the popular KDD Cup used a stacked ensemble having multiple logistic regression models in all of its ensembling layers[2].

Regularization is used to prevent overfitting. To apply L2 regularization, the logistic regression formula is extended to penalize high-magnitude coefficients by adding the penalty term in Equation 5 to the log loss function.

$$\frac{1}{C} \sum_i \beta_i^2 \qquad (5)$$

Here, $C$ is the inverse of the regularization strength. For highly collinear (and therefore partially interchangeable) features, the consequence of regularization is that the coefficient values will be more stable, but not necessarily more accurate in a descriptive sense. The value of $C$ is optimized using the validation set.

L2 regularization has been shown to improve performance for medium-term forecasts of extreme precipitation (Herman and Schumacher 2018).

### 3.1.2 Automated Machine Learning

Automated machine learning is an evolving field of research which aims to automate all aspects of machine learning including feature engineering, although success has mostly been achieved on model selection and hyperparameter optimization. One of the most popular automated machine learning suites is TPOT, which uses a genetic algorithm to find an optimal `sklearn` pipeline (Le, Fu, and Moore 2020) . Genetic algorithms imitate the process of natural selection. A pool of individuals is initialized and evaluated according to some fitness function. Then, the pool of individuals is updated by repeatedly combining two individuals, which are said to mate, and removing the unfit individuals from the pool. Fit individuals are more likely to produce offspring. One cycle of reproducing and removing unfit individuals is called a generation.

IN TPOT, each individual in the population is a machine learning pipeline. To produce offspring, two pipelines must share at least one primitive: this can be any input, such as a shared hyperparameter or a data input (Gijsbers, Vanschoren, and Olson 2018). In theory, TPOT could produce pipelines of arbitrary depth. In practice, TPOT tends to generate pipelines consisting of a single machine learning model without any pre- or post-processing steps, as automated preprocessing is still in its infancy. Nonetheless, it allows searching over a much larger model and hyperparameter grid than in a full grid search, being less computationally intensive. Model selection and hyperparameter optimization are performed simultaneously. TPOT has been shown to outperform methods based on random grid search (Balaji and Allen 2018)[3]. Like in random grid search, because the entire grid is not explored, there is no guarantee that the best-performing model and hyperparameters is tested.

---

[2]https://www.slideshare.net/jeongyoonlee/winning-data-science-competitions-74391113

TPOT was run for 5 generations, using a pool of 50 individuals. Two-fold cross-validation was employed for determining fitness. In each fold, random sample of 5% of 2015 and 2016 data for training and a random sample of 50% of 2015 and 2016 data for determining fitness. Since this process was computationally expensive, this process could only be run once. A lead time of 12 hours and a threshold of 1 mm were used. Even when subsampling only 5% of the data, some `sklearn` models included in the default pipeline grid configuration could not be fit within 5 minutes. Because of this, a restricted model and hyperparameter grid was created based on what TPOT developers call the "light" configuration, which only includes models which are relatively computationally inexpensive, such as linear models and nearest-neighbor classifiers. The light configuration was extended by adding the default hyperparameter grids for random forests and XGBoost. The full model/hyperparameter grid is provided in Appendix C.

One concern is that selecting a model based on the performance on random 2015 and 2016 data is not fair if the model is also tested on the same data: the model and hyperparameter selection process may itself overfit on the data. This does not appear to be the case. The best-performing pipeline, a random forest configuration, does not perform exceptionally well on 2015 and 2016 data when compared to the relative performance on 2017 data, nor does the selected pipeline perform better than other models.

### 3.1.3 Random Forests

The TPOT pipeline performing best on the dataset used ($l = +12$ hours, $h = 1$ mm) was a random forest configuration without any additional pre- or post-processing steps. This same configuration was also evaluated on other lead times and thresholds.

Random forests are an ensemble of decision trees (Breiman 2001). Random forests are able to model more complex interactions between predictands. For example, a logistic regression model is unlikely to have any use for the lat/lon data included, unless one of the ensemble members has a bias that is linearly related to coordinate position (e.g. one forecast consistently underestimates precipitation in the west of the Netherlands). A single decision tree in the forest, however, would in theory be able to learn that one forecast performs better in some regions of the country, and a second forecast performs better in others. Allowing for this type of freedom risks overfitting, hence why random forests, ensembling a large number of decision trees trained on different samples of the training set, typically perform better than single decision trees.

---

[3] After performing all experiments, I discovered that Zöller and Huber (2019) found no difference between random grid search and combined algorithm selection and state-of-the-art hyperparameter optimization methods, including TPOT

## 3.2 Neural Networks

### 3.2.1 Feedforward neural networks

Neural networks are non-linear models. With sufficient hidden layers and a non-linear activation function, they can approximate any continuous function (Cybenko 1989). The base unit of a neural network is a neuron, which takes $m$ true input signals $x_1$ to $x_m$, as well as a bias $x_0$. This bias is a constant, typically set to 1. $w_0$ to $w_m$ are learned weights. The output of the neuron is given in Equation 6.

$$o(w, x) = \varphi\left(\sum_{j=0}^{m+1} w_j x_j\right) = \varphi(w^T x) \qquad (6)$$

Here, $\varphi$ is the activation function, which is a differentiable scalar function. If $\varphi$ is the identity function, the neuron output is equivalent to that of linear regression when using $w_0$ as intercept and $w_1$ to $w_m$ as coefficients. If $\varphi$ is the sigmoid function, the output is equivalent to that of logistic regression, the sigmoid being the inverse of the logit function described in the section on logistic regression.

In a multi-layer feedforward neural network, the output of one group of neurons is used as the input of another. In a sequential neural network, the first group of neurons takes a data sample as input, and each subsequent group takes the output of the neurons of the group before it as input (Figure 4). Each group of neurons is called a layer. In a densely-connected neural network, all possible connections between two subsequent layers are made.

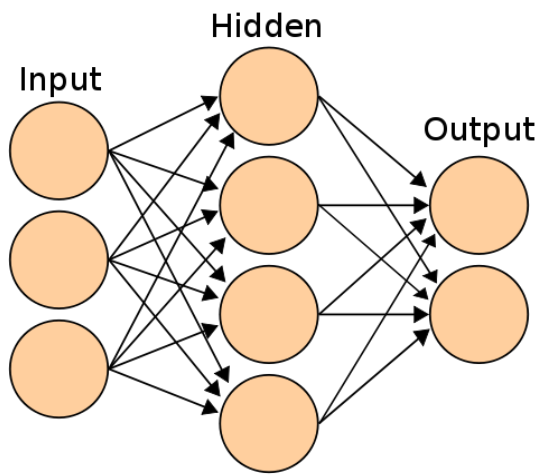

Figure 4: Densely-connected neural network with a single hidden layer

While the above globally describes how neural networks form predictions given already learned weights and data, it does not describe how weights are learned. In the simplest case, weights are iteratively updated for each labeled sample. In binary problems, the target label $t$ is set to 1 for positive samples and to 0 for negative samples.

For a single neuron, the update rule is known as the delta rule. First, the local gradient is defined in Equation 7.

$$\delta = (\varphi(w^T x) - t)\varphi'(w^T x) \tag{7}$$

The delta rule (Equation 8) then describes how the weights of the single neuron are updated to slightly descend that gradient.

$$\Delta w_i = -\alpha \delta x_i \tag{8}$$

In Equation 8, $\alpha$ is a small constant known as the learning rate.

In multi-layer feedforward neural networks, the output neurons, having direct access to the error signal, are also updated using the delta rule above.

The weights of the non-output neurons, also known as hidden neurons, are then recursively updated in a process called backpropagation (Equation 9).

$$\delta_j = \varphi'(w_j^T x) \sum_{k \in K} w_{kj} \delta_k \tag{9}$$

In autoref{backpropeq}, $\delta_j$ is the local gradient of neuron $j$, $K$ is the set of neurons receiving a signal from neuron $j$, and $w_k j$ is the input weight in neuron $k$ for the signal coming from neuron $j$.

Using the same weight notation:

$$\Delta w_i j = -\alpha o_i \delta_j \tag{10}$$

In autoref{lastnneq}, $o_i$ is the output signal from neuron $i$, already transformed by the activation function.

For a detailed derivation of backpropagation, see Bishop (2006).

### 3.2.2 Convolutional Neural Networks

Convolutional neural networks (CNNs) are a variant of feedforward neural networks which are commonly used in image-to-image learning. CNNs have been used as end-to-end models in short-term precipitation forecasting (Shi et al. 2017).

Tasks where each sample has a notion of dimensionality, such as image classification and text classification, often rely on features that are invariant to translation. For instance, a binary fork/spoon image classifier should classify a spoon as a spoon, whether it appears in the top left corner of the image or in the bottom right. To some degree, this could also be true for post-processing precipitation forecasts. If, to take an unlikely example, one forecast is generally overconfident about the exact location where precipitation will fall, blurring the forecast image could lead to better results. Blurring, like many local image transformations, can be performed using convolution (Equation 11).

$$g(x,y) = \omega * f(x,y) = \sum_{dx=-a}^{a} \sum_{dy=-b}^{b} \omega(dx, dy) f(x+dx, y+dy) \tag{11}$$

Here, $f$ is the original image and $\omega$ is small matrix called the filter kernel. $\omega$ is indexed such that the x- and y-coordinates run from $-a$ to $a$ and from $-b$ to $b$, respectively, with $(0, 0)$ at the center of the kernel. As such, this 3x3 kernel is used for the identity operation:

$$\begin{bmatrix} 0 & 0 & 0 \\ 0 & 1 & 0 \\ 0 & 0 & 0 \end{bmatrix}$$

Blurring can be done by averaging over all neighbouring pixels in a 3x3 box:

$$\begin{bmatrix} \frac{1}{9} & \frac{1}{9} & \frac{1}{9} \\ \frac{1}{9} & \frac{1}{9} & \frac{1}{9} \\ \frac{1}{9} & \frac{1}{9} & \frac{1}{9} \end{bmatrix}$$

Convolution operates over neighboring pixels. Because of this, the operation is poorly defined for edge pixels. Valid convolution simply does not return results for edge pixels, meaning that the output image is smaller than the input image for any kernel larger than 1x1. In our final architecture, we use convolution with zero-padding, where each input value out of bounds is assumed to be 0.

A single-layer convolutional neural network is nothing more than a convolution operation followed by a per-pixel activation function. Like the weights in other neural networks, the values in $\omega$, known as the kernel weights, can be trained using gradient descent. Note that to learn the blur operation, the neural network would only need to learn 9 weights plus one bias, regardless of the number of pixels in the input image. By comparison, a single-layer densely-connected network for images of size $h \times w$ with an output of the same size would have $h^2 w^2$ weights, one connection for each pair of input and output pixels.

When a CNN is trained on multiple 2D features, like input forecasts, these are concatenated into a 3D matrix, and a 3D kernel is applied. The kernel size along the new axis is equal to the number of features, and the kernel does not convolve along the third axis. This is so common that, if the context allows for it, a 3x3 kernel should be taken to refer to a 3x3x$c$ kernel, where $c$ is the number of features, which is also referred to as the number of channels. As the 3D kernel only truly convolves along two dimensions, this operation is still called a 2D convolution.

By applying multiple layers of convolution, more complex operations can be performed. The kernel weights of hidden CNN layers can, too, be trained using backpropagation. However, a 2D convolution layer as described above only outputs a single 2D image, meaning that the first layer would reduce the input from $h \times w \times c$ to $h \times w$, forcing heavy compression of information in the first layer of the network. To prevent heavy compression, multiple 2D convolutions are applied to the input, and their 2D outputs are concatenated into a 3D matrix, which is used as input for the second layer. Because this trick is usually repeated in every layer, a set of $c_o$ 2D convolutions taking a $w \times h \times c_i$ matrix and outputting a $w \times c_o$ matrix is still considered a single 2D convolutional layer.

Having multiple output channels per convolutional layer does not just prevent compression. Increasing the number of output channels to a number larger than the number of input channels in the first part of the network allows the network to learn a wide range of intermediary features. For instance, in many CNN architectures developed for image classification tasks, the first CNN layer learns to function as an edge detector (e.g. Zeiler and Fergus 2014), where each output channel fires upon seeing an edge at a specific angle. As shown in the results, no human-interpretable features were found in the first layer of our models (Figure 12).

Convolutional layers are limited to local information. As such, after a limited number of convolutions, each output pixel only depends on input pixels in the same neighborhood. This makes it impossible for the network to correct errors of a scale larger than that neighborhood. Possible ways to increase that scale are to increase the number of layers and to increase the kernel size, but this is only computationally feasible to a certain point. In addition, increased complexity brings about risk of overfitting.

An alternative method that performs well in practice is pooling (e.g. in Badrinarayanan, Kendall, and Cipolla 2017; Ronneberger, Fischer, and Brox 2015). Pooling is an operation where each channel is resized to a lower resolution. This requires combining neighboring pixels. Max-pooling is most commonly used, using the maximum of all source pixels as the output (Figure 5). Layers following max-pooling operations will operate on a larger neighborhood than the layers before, even if the kernel size remains the same. Max-pooling is usually combined with an increase in channel depth, allowing the layer to represent more information per pixel, each pixel now spanning a larger area.

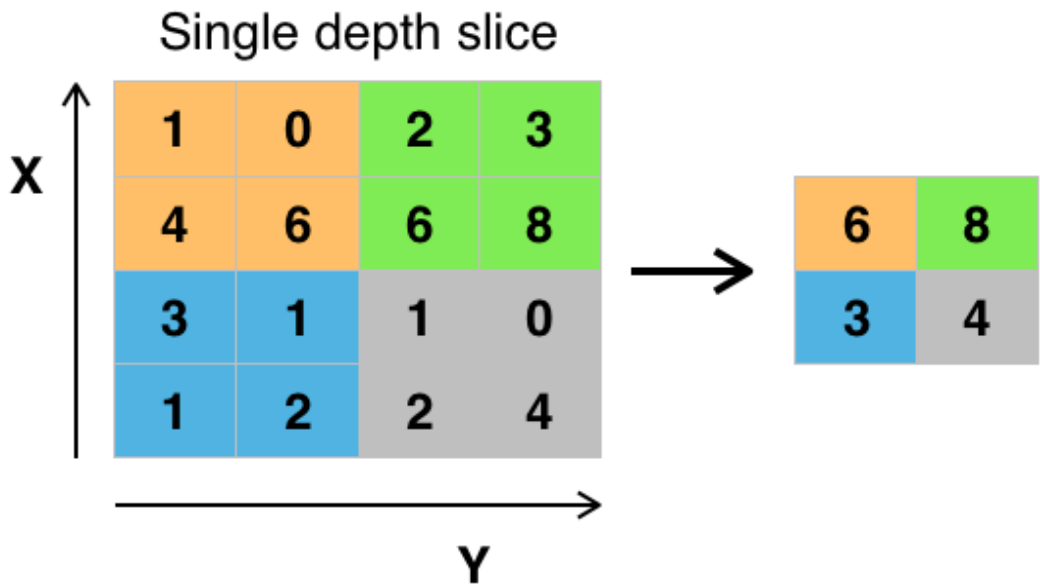

Figure 5: 2x2 max-pooling on a single channel

The result of repeated max-pooling is an image much smaller than the input. This is problematic if the output should be of the same resolution as the input image, as is wanted in this thesis. One could use one or more densely connected layers to transform the low-resolution representation, which has a high channel depth, to the high-resolution output representation, which usually has a comparably low channel depth. Alternatively, the process of max-pooling can be reversed in the second half of the network, resizing each channel to increase the resolution. The first section of the network, where the resolution is decreased and the channel depth is increased, is referred to the encoder; the second section that increases the resolution and decreases the channel depth is referred to as the decoder[4]. Recurrent convolutional encoder-decoder architectures are already commonly used in short-term precipitation forecasting (e.g. Shi et al. 2015). Convolutional encoder-decoders are also commonly used in tasks such as image segmentation (e.g. Ronneberger, Fischer, and Brox 2015).

Table 4: Hyperparameters of interest in the SegNet-based architecture which were not included in any grid search. *Values explored* were compared based on validation set performance while still developing the model, usually on a single lead time and a threshold of .5 mm and/or 1 mm. *Final value* denotes the value used in the final hyperparameter optimization.

| Hyperparameter | Values explored | Final value | Description |
| --- | --- | --- | --- |
| $\alpha$ | 1e-5, 1e-4, 1e-3, 1e-2, 1 | 0.01 | Learning rate |
| DENSE | True, False | False | Whether to use dense layers rather than a fully-convolutional decoder |
| REWEIGHT | True, False | False | Whether to use balanced class reweighting |
| X_LAST | True, False | True | Whether to reintroduce all features before final convolution |
| REMOVE_MIDDLE | True, False | False | Whether to remove 10 of the middle layers of Segnet for a comparatively shallow architecture |
| SKIP | True, False | False | Whether to use skip connections |
| UNPOOL | True, False | True | Whether to use unpooling layers rather than upsampling in the decoder |
| ODD_POOL | True, False | True | Whether to modify SegNet by pooling after the first and fifth layer and convolving in the middle |
| KERNEL | 2, 3, 4 | 3 | Kernel size |
| FILTER_MUL | 1, 2 | 2 | Multiplication factor for the no. of filters |
| SELECTED | Detailed below | Detailed below | The selected feature set |
| $\phi$ | | ReLu | Activation function |
| LATE_CONV | True, False | False | Whether to perform two 1x1 convolutions after reintroducing the features rather than one |
| PATIENCE | 20, 40, 400 | 20 | The number of epochs without improvement before stopping |
| LARGE_KERNEL | 3, 4, 5, 6 | 3 | Kernel size in the first few layers |
| REDUCE_LR | True, False | True | Whether to reduce the learning rate on a plateau |
| OPTIMIZER | Adam, SGD with/without momentum | SGD (momentum = 0.9) | The optimizer used for the CNN |
| BATCH_SIZE | 5, 16, 32 | 32 | The batch size of the CNN |

---

[4]I use a broad definition of encoder-decoders, where the two parts are not necessarily decoupled and there is not necessarily one "correct" latent space

| Hyperparameter | Values explored | Final value | Description |
| --- | --- | --- | --- |
| LOSS | log loss, MSE (Brier score) | log loss | The loss function used |

It was infeasible to perform a full grid search over all hyperparameters of interest shown in Table 4. As such, their values were selected based on a comparison of validation set performance during development. Only a subset of lead times and thresholds was used for these decisions. The choice of lead times and thresholds was non-systemic; it would usually just be the lead times and thresholds I was experimenting with at the time.

As stated in Table 1, there are 38028 output pixels for which predictions are made. As such, densely connected output layers were computationally expensive. In the dense networks tested, the dense layers accounted for the majority of parameters. Given that the dense networks performed slightly worse in initial tests, that many performant encoder-decoders for short-term forecasting are fully convolutional, and that there were already a large number of hyperparameters to be optimized, experiments with densely connected layers were aborted in early stages. An additional advantage of this choice is that since all encoder-decoders tested are fully convolutional, they can be trained on areas of any size without an increase in parameters.

The final fully-convolutional architecture was based on SegNet (Badrinarayanan, Kendall, and Cipolla 2017). U-net (Ronneberger, Fischer, and Brox 2015) was also explored, and there were attempts to combine the two architectures in various ways. As in SegNet, the ReLu activation function and batch normalization were used for every layer.

U-net and SegNet are encoder-decoders designed for image segmentation, but they are both quite generalist. U-net has notably already been used as an end-to-end system for short-term precipitation forecasting (Agrawal et al. 2019). The number of layers and the number of filters for each layer, as well as the distribution of pooling layers, were adapted from SegNet and U-net. Like SegNet and U-net, the postprocessing model performs 2x2 max-pooling after every second convolutional layer. Also like SegNet and U-net, the number of channels is doubled after each pooling step.

U-net and SegNet are both minimal extensions of the vanilla encoder-decoders detailed above. As such, they are very similar to the non-recurrent part of the end-to-end nowcasting architectures described in the introduction.

U-net is a vanilla encoder-decoder save for skip connections. Skip connections reintroduce the output of the $n$th encoding layer as the input of the $n$th last hidden decoding layer (Figure 7).

SegNet is even closer to a vanilla encoder-decoder, having an equally low number of parameters and being as swift to train. The novelty is that it uses unpooling layers (Figure 6) rather than simple upsampling or a learned upsampling method such as transposed convolution.

SegNet, having only five 2x2 pooling operations, does not reduce images to a 1x1 representation in the middle layer. Like for image segmentation, this is not necessarily a bad thing for forecast postprocessing: the CNN will likely be best at correcting errors at a smaller scale.

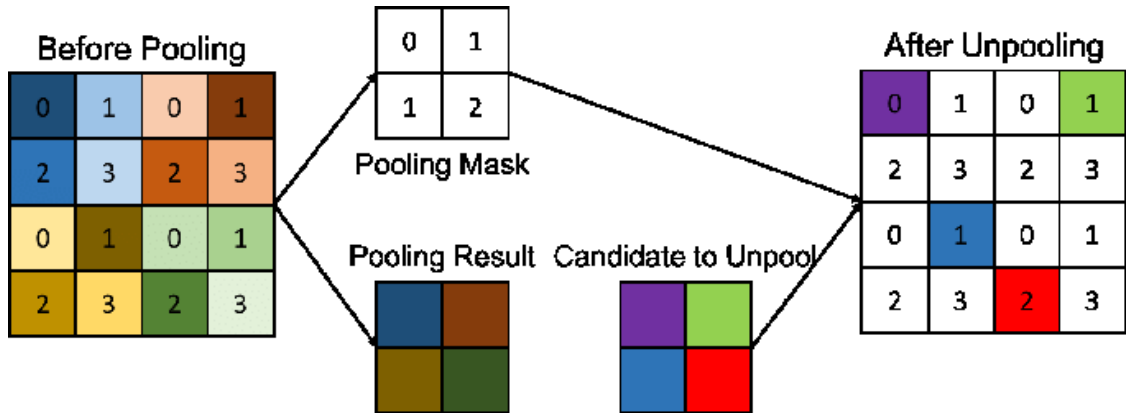

Figure 6: 2x2 max-pooling and unpooling operations. Darker squares denote higher values.

In the early stages of development, both SegNet and U-net were trained and tested with very minimal alterations, only applying the masking and the changes to output detailed below. A +12 lead time and a 0.5 mm threshold were used. SegNet was found to converge faster and perform slightly better on the validation set than U-net at any epoch. Skip connections and simple upsampling were then transferred to SegNet and unpooling was implemented in U-net to differentiate between the possible causes for SegNet's superiority. Skip connections did not make SegNet more performant. Removing unpooling from SegNet did slightly decrease performance.

After initial testing, I decided to base my architecture on SegNet because it was faster to train, had a smaller number of parameters and performed slightly better on a small sample of lead times and thresholds. Skip connections were not included, as they increased the number of parameters and made no difference in development testing. Unpooling was still used even though the difference in validation performance was quite small and could be attributed to chance, because it did not require additional parameters to be learnt, it hardly requires any additional computation, and there is little chance that unpooling would have a negative effect.

Regridded forecasts were cropped to 384 by 384 pixels, corresponding roughly to an area spanning 384 by 384 km. 384 was chosen because it factorizes to 3 x $2^7$, allowing for straightforward 2x2 pooling and unpooling. The forecasts were cropped such that input forecasts for at least 30 km outside the output forecast area were included, allowing input forecasts for the sea and neighboring border areas to influence the output forecasts.

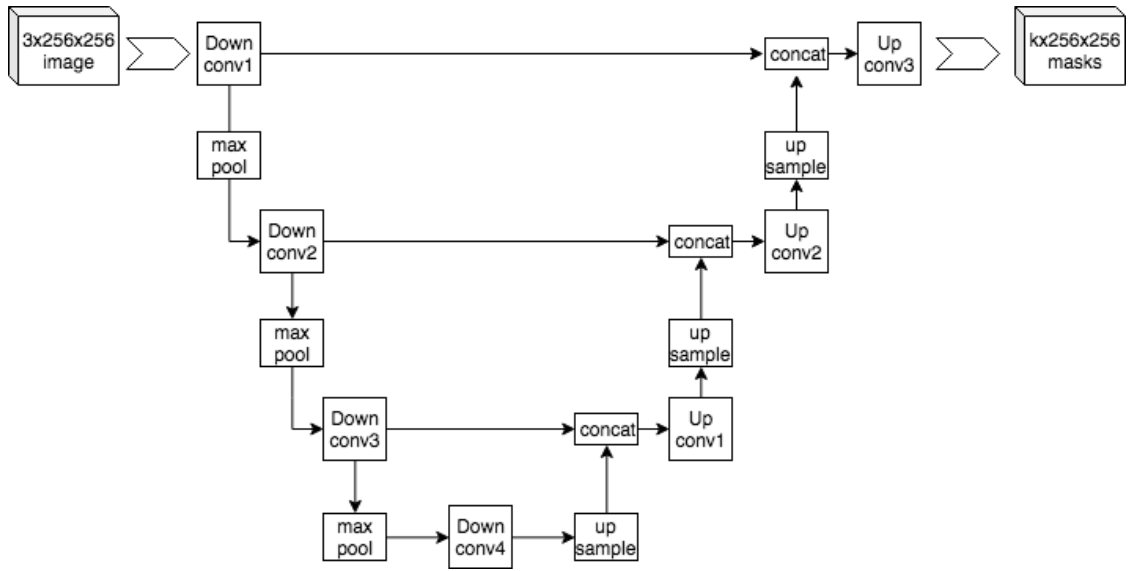

Figure 7: An example architecture with skip connections

Required changes to U-net and SegNet were reducing the output channels to one—the task is modeled as binary classification—and masking the output to the valid observations (Figure 2), since a CNN usually outputs a rectangular image. Initially, the masking was implemented partially outside the neural network. Before outputting this image, it was multiplied by the observation mask, guaranteeing that forecasts for unobserved pixels would only ever be 0. The labels for unobserved pixels were also set to 0. As such, these pixels should never contribute to the loss. However, after poor initial performance of the CNN, this was replaced by a functionally equivalent but more straightforward operation included in Tensorflow, `tf.boolean_mask`, followed by flattening the output. This change had no influence on performance. In the Keras summary of the network in Appendix D, the masking operation occurs in the Lambda layer.

Veldkamp (2020) found that a quantile regression forest directly postprocessing wind speed forecasts performed much worse than a quantile regression forest stacked on top of logistic regression. Although the same was not found for CNNs, the approach did inspire a modification to SegNet. In Veldkamp's stacked model, linear regression was fit on the predicted wind speed. The CNN was trained to predict the error of the linear regression. The final forecast added the two predictions.

Since the tasks in this thesis are modeled as a classification problem, the approach cannot be used directly. A merger of a simple linear model and neural networks does, however, seem like a sensible idea, given that simple linear models perform very well by themselves in meteorological post-processing. This was implemented by reintroducing all features just before the final 1x1 convolution. This can also be seen as a skip connection from the input to the final 1x1 convolution. It is quite similar to a stacked classifier. If one would train the weights of one pathway while freezing the other, this would truly be a stacked model. Seeing no reason not to train the pathways at the same time, no weights were frozen.

Introducing geographical coordinates (features `xdim` and `ydim`) before the final 1x1 convolution has already been shown to be helpful for CNN architectures (Liu et al. 2018).

Early stopping was used for all CNNs. The number of epochs without improvement after which training stopped was determined by the `PATIENCE` hyperparameter. A final value of 20 was used. Early stopping was performed based on the model's Brier score on the validation set. After training, the weights from the epoch with the best validation Brier score were restored. The Brier score is an evaluation metric for probabilistic forecasts described in the *Results* section.

Higher values for `PATIENCE` were tested during development (Table 4). Interestingly, the CNN did not overfit at any point during development. As a sanity check, the model did was trained on 40 input images; this did result on overfitting, as expected. In runs using all training data, training loss and validation loss simply reached a plateau (Figure 22). In an attempt to break the plateau, the learning rate was dynamically reduced by a factor of 10 if the validation Brier score did not improve for 10 epochs. This did not cause any improvement.

During development, classes were reweighted for both the CNNs and the Keras-based logistic regression using the "balanced" heuristic formulated by King and Zeng (2001). Here, the validation Brier score was not used for early stopping, since the Brier score was not affected by the custom reweighting function I implemented for Keras[5]. Instead, the validation loss was used. Models fit on reweighted classes are biased and do not do well when uncalibrated, so I looked at test set performance for $h = 0.5$ mm, $l = +12$ to see if the reweighting was useful. No increase in performance was found.

Most models for short-term precipitation forecasting use a smaller number of filters, although I was unable to find research showing a one-to-one comparison of different filter counts. In line with models for short-term precipitation forecasting, the number of filters in each SegNet layer was initially reduced by a factor of 8, or, if `FILTER_MUL` was set to 2, by a factor of 4. As such, the number of filters in the outermost layers was 8 or 16.

After the CNN failed to beat `sklearn` logistic regression on the validation sets of various subexperiments throughout development, I wanted to verify that the difference was not the result of an implementation detail or a mistake in either postprecessing pipeline. In addition, I wanted to see if adding early stopping helped fitting logistic regression even more.

To use the same pipeline, optimizer and early stopping rules as the CNN, per-pixel logistic regression was implemented in Keras as a CNN with a single 1x1 layer and a sigmoid activation function. In the results, the Keras version is labeled as *LR with early stopping*.

During development, I also experimented with adding a single hidden 1x1 convolutional layer with 8 output channels to the logistic regression model described above, which would make it equivalent to a multi-layer perceptron with one hidden layer. This did not increase validation performance.

# 4 Results

Model performance is evaluated using the Brier score, a proper scoring rule, equivalent to the mean squared error between the predicted probabilities and numerical target labels. Here, labels for positive samples are set to 1 and a labels for negative samples are set to 0.

The Brier score is by far the most commonly used measure for probabilistic binary classification problems in meteorology (Wilks 2019, 331). Even though it is called a score, lower values are better. The Brier skill score (BSS) is a comparative score aiming to describe a model's Brier score relative to a baseline model (Equation 12).

---

[5]This function reweighted the classes on the level of atomic samples

$$BSS = 1 - \frac{brier_{model}}{brier_{baseline}} \qquad (12)$$

In this thesis, the baseline for Brier skill scores is always per-pixel climatology, which is defined as the fraction of positive samples for each pixel, or the mean of the numerical label vector[6]. As such, the BSS is 0 if the model's Brier score is equivalent to climatology. 1 is a perfect BSS.

All results pertain to both winter and summer, unless stated otherwise.

Table 5: Summary of output predictions

| Name | Description |
| --- | --- |
| gefs | Isotonic regression applied to the GEFS ensemble mean |
| harmonie | Isotonic regression applied to the HARMONIE forecast |
| climate | Climatology; per-pixel baseline probabilities |
| LR with(out) early stopping | Calibrated logistic regression with(out) early stopping |
| CNN | Calibrated CNN |

Although calibrating the raw ensemble forecast would be more intuitive than calibrating the ensemble mean for a probabilistic forecast, the calibrated ensemble mean consistently outperformed the calibrated raw ensemble forecast on the validation set.

All test set forecasts are properly calibrated for lower predicted probabilities at the lowest threshold (Figure 8). For higher probabilities, especially the GEFS ensemble mean calibrated using isotonic regression seems poorly calibrated. Given that isotonic regression is a monotonic calibration method, Figure 8 suggests that there is a non-linear relationship between the GEFS ensemble mean and precipitation exceeding the 0.5 threshold. However, only less than 0.1% of atomic observations has a calibrated 12-hour GEFS forecast of 66% or higher (Figure 9). This is presumably because GEFS, with its low spatial and temporal resolution, is relatively imprecise. As such, the relation between calibrated GEFS forecasts and precipitation is unlikely to actually inverse for larger precipitation amounts, and is rather a fluke resulting from the small effective sample size. Nonetheless, there is no reason to believe the high-probability forecasts are well-calibrated.

For 2 mm thresholds, calibration is unstable for forecasts over 30% (Figure 10; Figure 11). In the subexperiment shown, the CNN is better calibrated than the rest, but this is not generally true for 2 mm forecasts.

---

[6]The Brier score is equivalent to the mean squared error and the variance of a variable is equivalent to the mean squared error of a constant mean prediction for that variable. Because of this, the baseline performance is equal to the variance of the label vector.

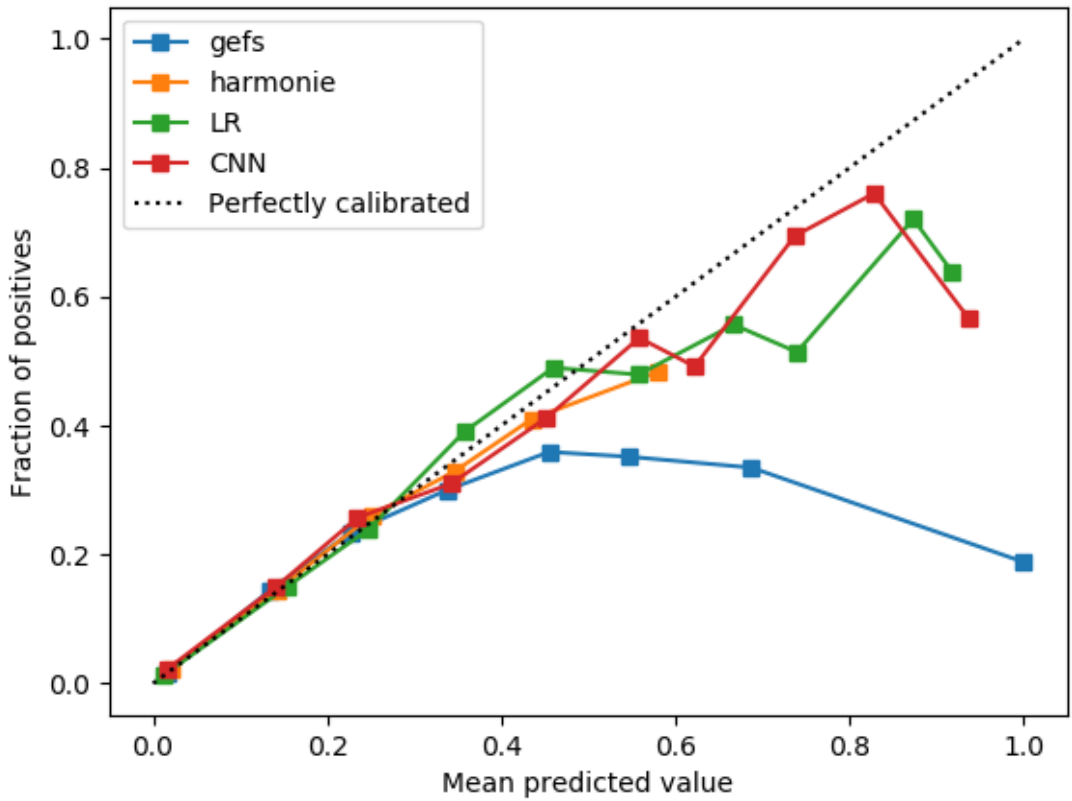

Figure 8: The calibration curve for the calibrated test set forecasts ($l = +12$, $h = 0.5$ mm)

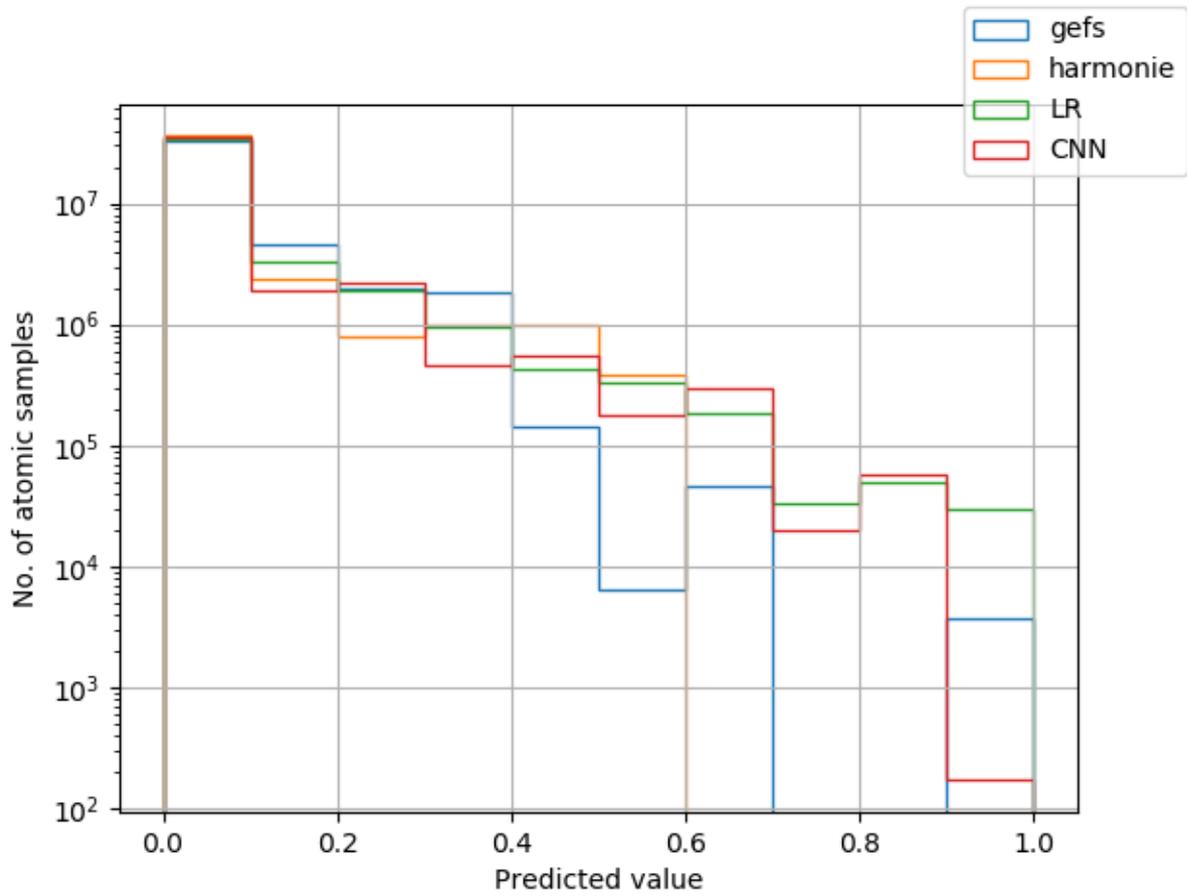

Figure 9: Frequencies of output predictions and calibrated input predictions ($l = +12$, $h = 0.5$ mm). Note that samples are correlated; effective sample size is lower.

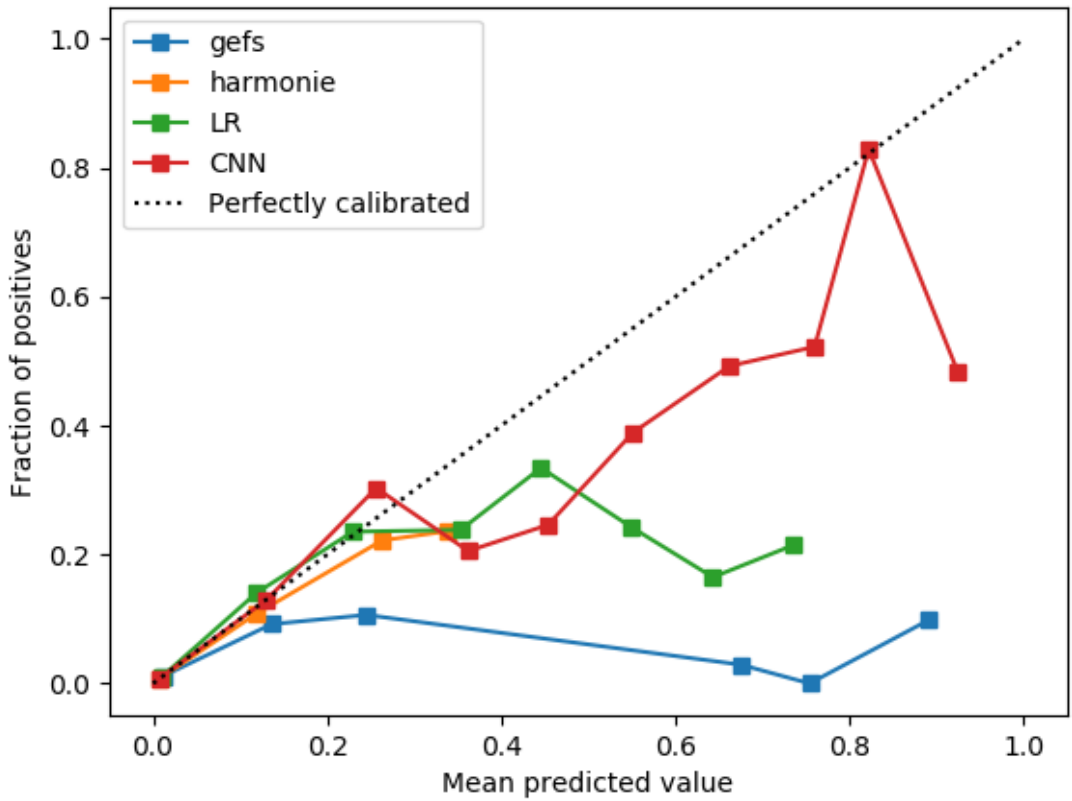

Figure 10: The calibration curve for the calibrated test set forecasts ($l = +12$, $h = 2$ mm)

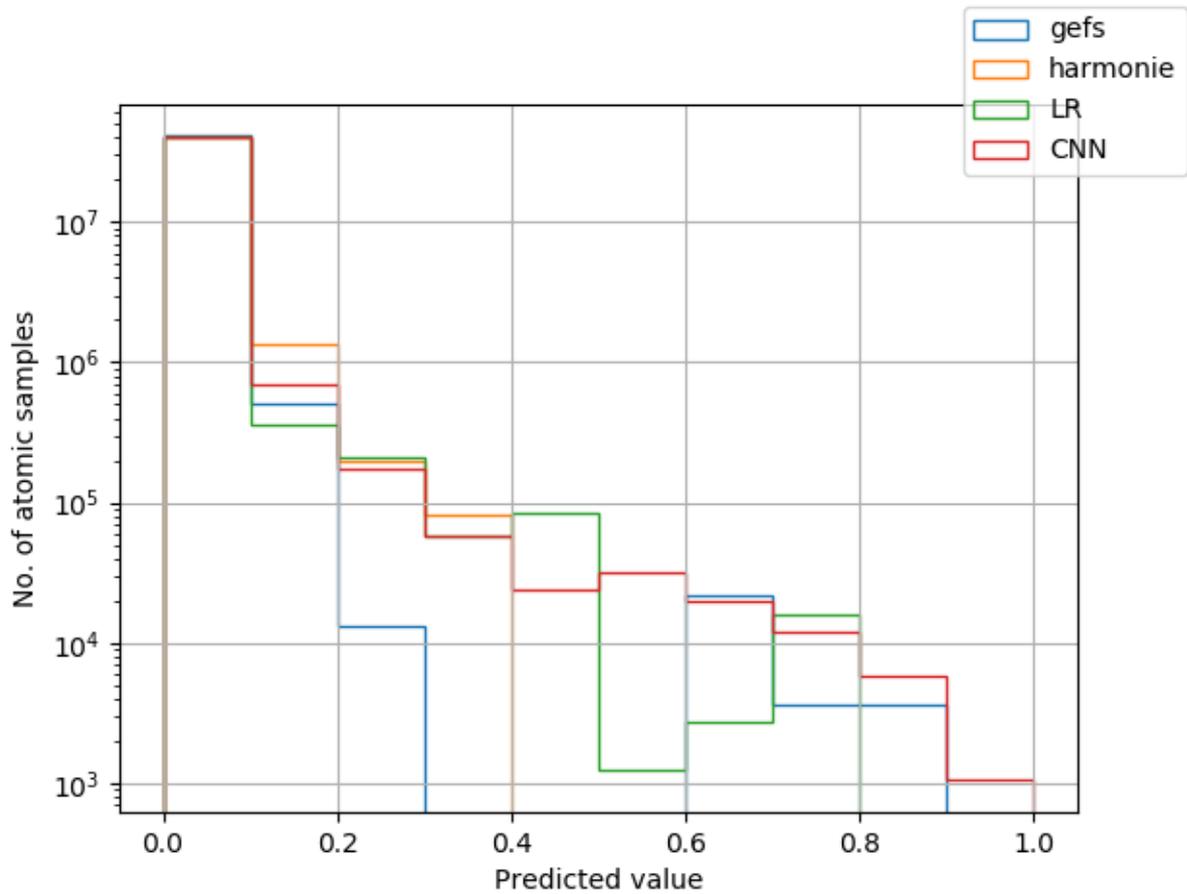

Figure 11: Frequencies of output predictions and calibrated input predictions ($l = +12$, $h = 2$ mm). Note that samples are correlated; effective sample size is lower.

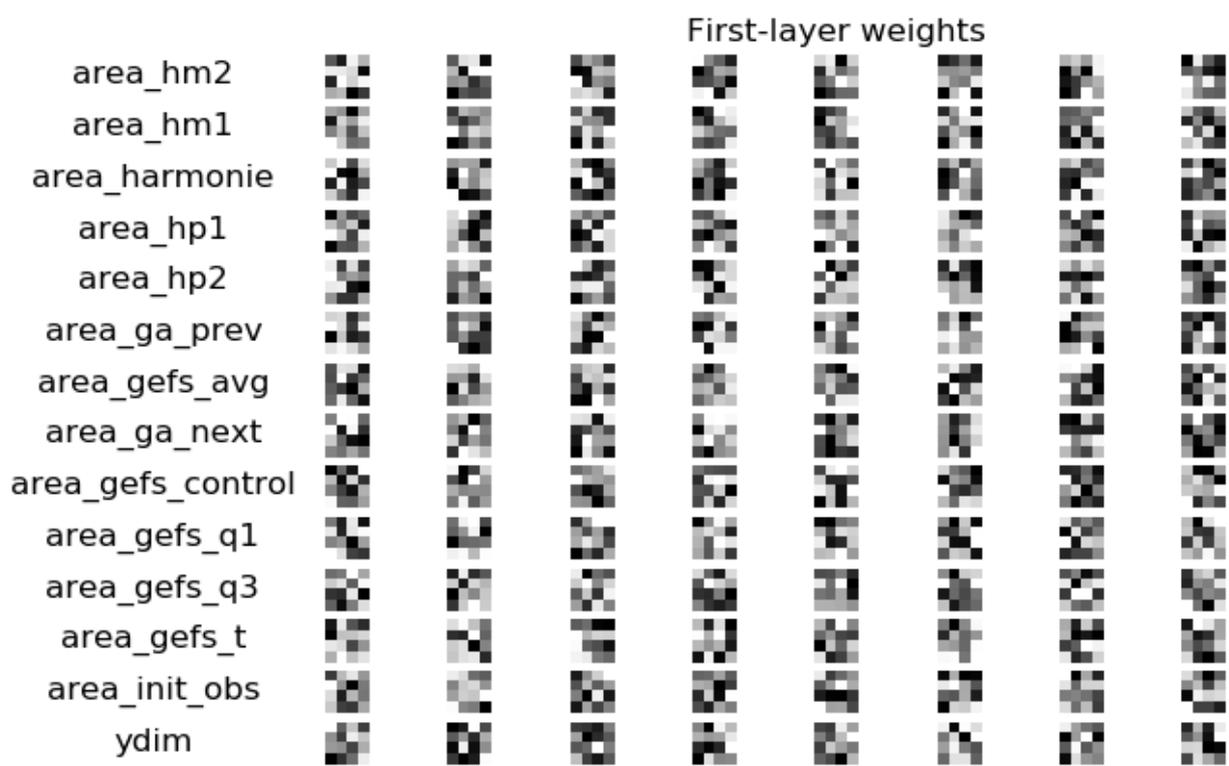

Figure 12: First layer weights for 4x4 kernel, 0.5mm, 24 hour lead time, winter only

The first-layer weights in Figure 12 do not show human-interpretable patterns found in picture classification tasks (as in e.g. Zeiler and Fergus 2014). I was unable to find work on short-term precipitation forecasting that included first-layer weights. It is not clear if performant short-term precipitation models do display human-interpretable first-layer behavior.

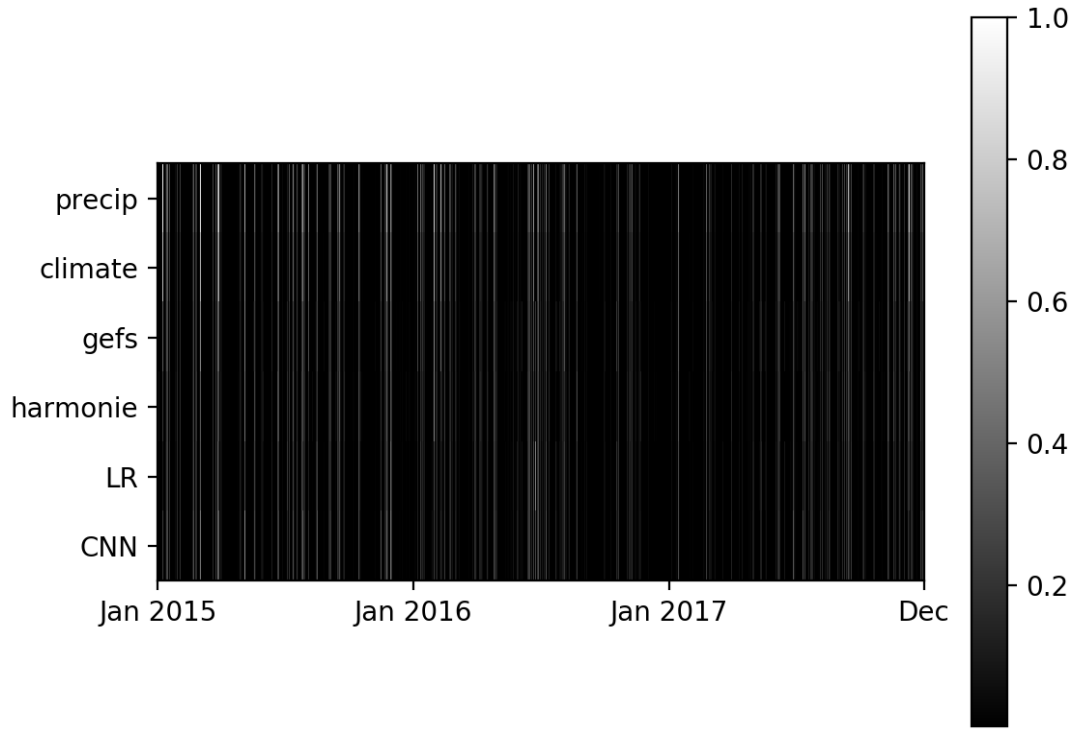

Figure 13: Top row shows the +12 precipitation cover for each day. The remaining rows are the daily Brier scores for each output forecast ($l = +12$, $h = 0.5$ mm, brighter is worse)

The days with most precipitation contribute the most to the composite Brier scores, and a large number of days has low precipitation probabilities both in the input forecasts and the output forecasts (Figure 13). This is to be expected, as the baseline probability for precipitation is quite low even for low thresholds (Table 3). Both logistic regression and CNNs are resilient to situations where one model forecasts poorly and the other does not.

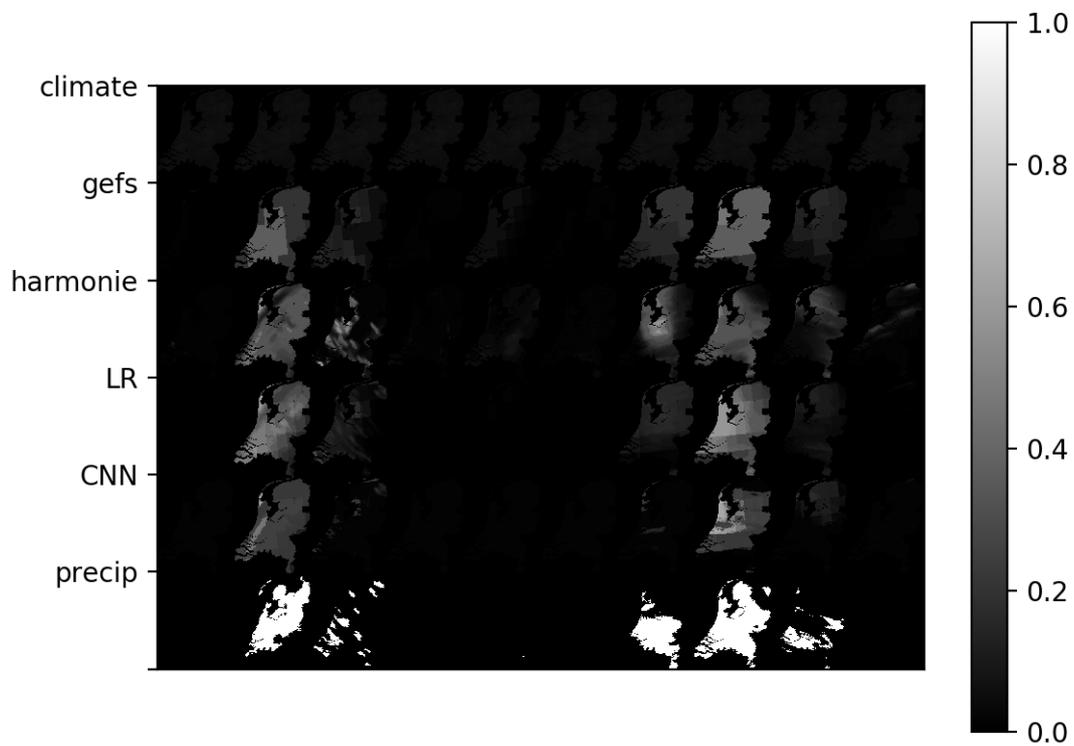

Figure 14: Output forecasts for 2015-11-22 to 2015-12-01, a rainy period ($l = +12$, $h = 0.5$ mm)

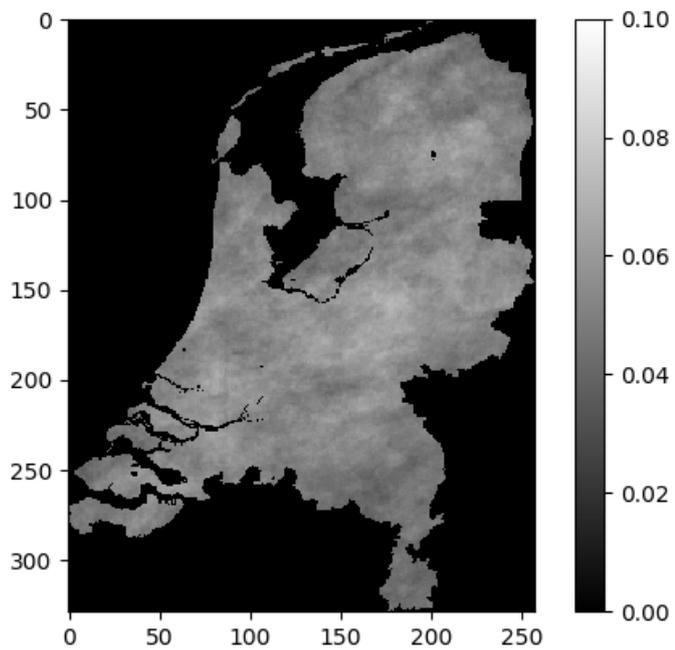

Figure 15: Brier scores for climatology predictions ($l = +12$, $h = 0.5$ mm, brighter is worse)

There are no striking differences between logistic regression and CNN forecasts (see Figure 14 for a representative example). In particular, none of the CNN forecasts I investigated appeared to have any spatial transformation of the input forecasts. The final forecasts seem to rely heavily on the short pathway between input and output rather than the encoder-decoder. The CNNs without this path tested during development (i.e. where the hyperparameter `X_LAST` was set to false), simply produced slightly blurrier versions of the input, performing even worse.

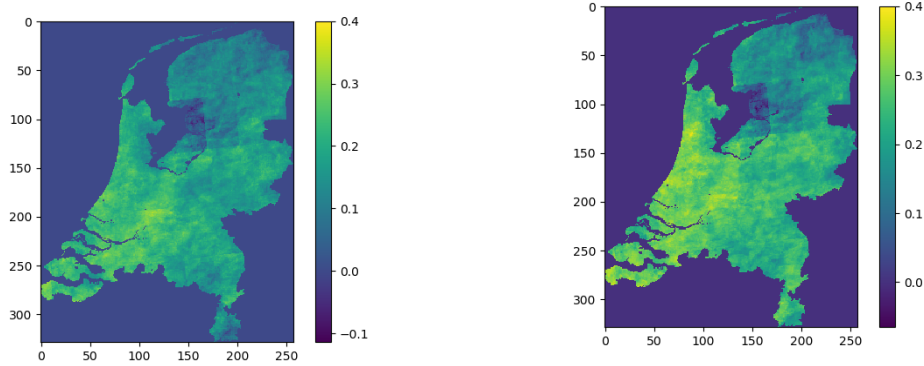

(a) Per-pixel CNN BSS ($l = +12$, $h = 0.5$ mm)

(b) Per-pixel logistic regression BSS ($l = +12$, $h = 0.5$ mm)

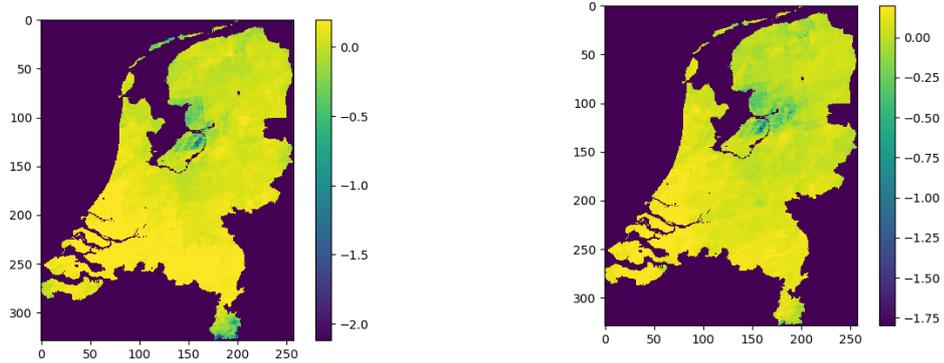

(c) Per-pixel CNN BSS ($l = +12$, $h = 2$ mm)

(d) Per-pixel logistic regression BSS ($l = +12$, $h = 2$ mm)

Figure 16: Per-pixel BSS (brighter is better). Reference brier scores are shown in Figure 15.

All models forecast poorly in the GEFS grid cell that's partially on the IJselmeer, partially east of it (the consistently dark square in Figure 16). This is the result of GEFS forecasting poorly in this region. It could be that GEFS is forecasting correctly for the lake itself, but the surrounding land mass has different weather patterns. This cannot be easily checked since the ground truth used only covers land mass.

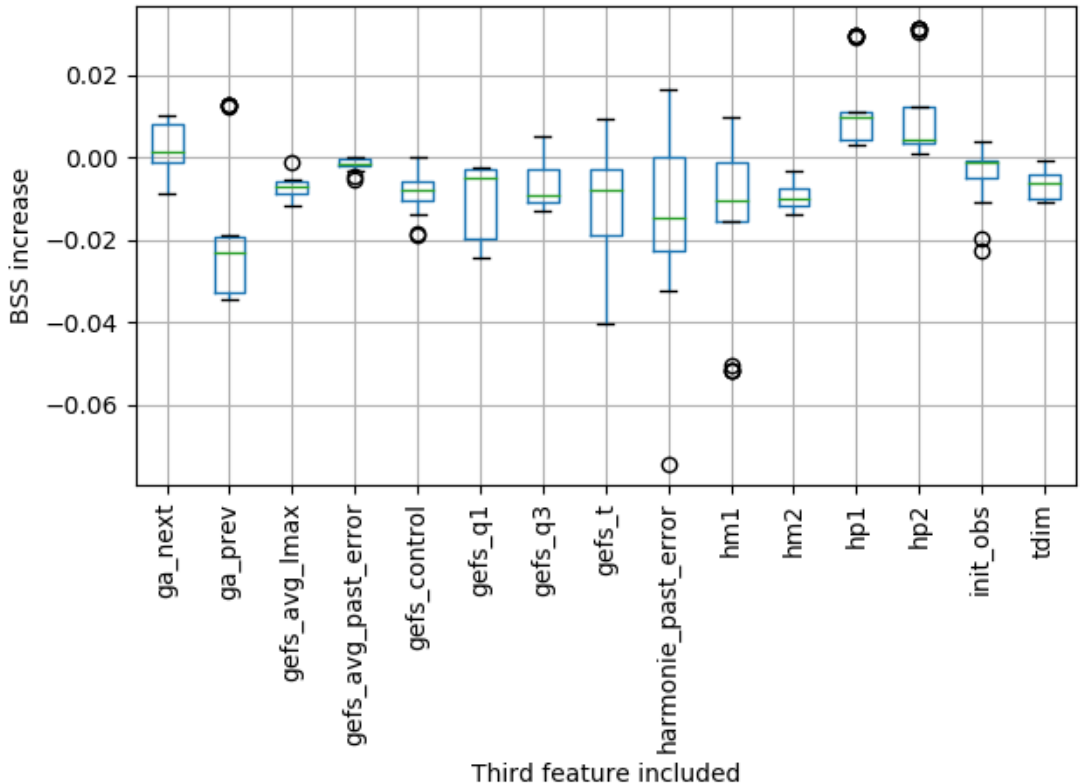

Figure 17: BSS increase from using a third feature in addition to the HARMONIE forecast and the GEFS ensemble mean (logistic regression, all hyperparameter configurations, all lead times, all thresholds).

For the ablation analysis, we considered forecasts based on the GEFS ensemble mean and the HARMONIE to be the reference forecast. The ensemble mean is a robust derived feature. It is consistently assigned a high permutation importance in meteorological post-processing studies (e.g. Eccel et al. 2007). Based on the correlation matrix (Appendix A), it seemed somewhat doubtful that other features derived from GEFS would add anything when the ensemble mean was already included. Although an ablation analysis by name, some results are reported as the BSS increase from adding a feature, not as the BSS decrease from removing a feature.

The addition of most features on top of the HARMONIE forecast and the GEFS ensemble mean do not increase logistic regression performance (Figure 17). `hp1` and `hp2`, the HARMONIE forecasts for the two hours after validation time, do contribute to performance, suggesting a systemic bias in the precipitation forecasts with respect to timing, or a systemic bias in the degree of the persistence of precipitation.

In all but one run in one subexperiment, including GEFS in addition the HARMONIE forecast improves the BSS of logistic regression, despite the low spatial temporal and spatial resolution in GEFS (Figure 18). The addition of HARMONIE to GEFS always improves the performance of logistic regression.

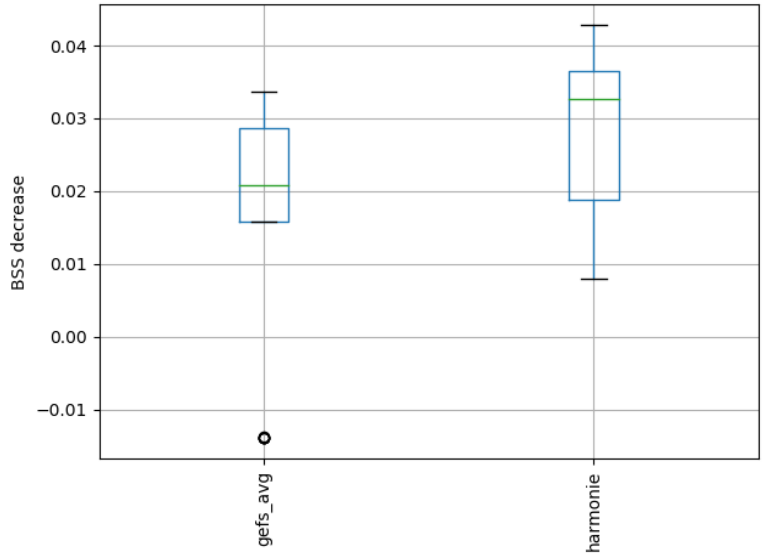

Figure 18: BSS decrease from removing a feature from the set of two reference features (LR, all lead times, all thresholds, all hyperparameter configurations)

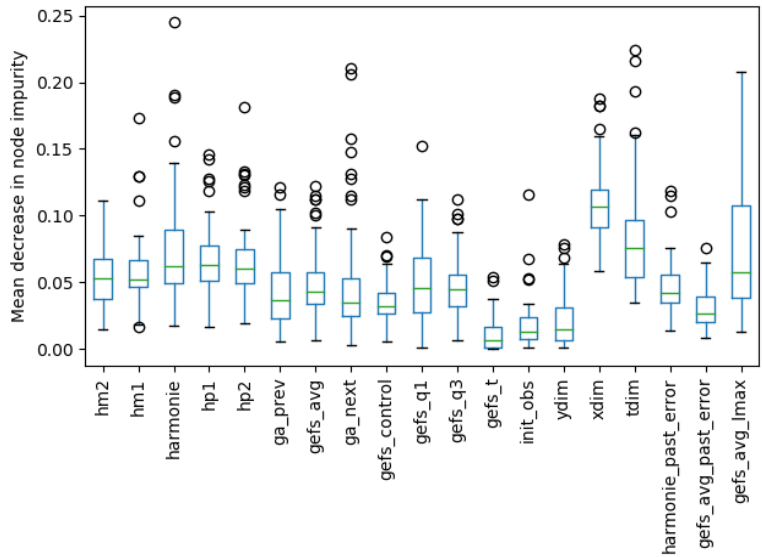

Figure 19: Random forest mean decrease in impurity (all leads, all thresholds, all folds). Should **not** be taken to be the true feature importance.

Many features derived from the GEFS forecast are considered equally important based on the mean decrease in impurity (Figure 19). However, removing the first quartile, third quartile and control forecast from the feature set does not seem to have any influence on the Brier skill score (Figure 20). This underscores the fact that the marginal benefit of including a feature cannot be determined using mean decrease in impurity. This is discussed in further detail in the next section.

`xdim` is considered the most important feature, but this likely only so because there are no redundant features `xdim` has to share its importance score with, as is the case for the features based on HARMONIE and GEFS forecasts. Nonetheless, it is interesting to see that the random forest is capable of using geographical information which is itself very weakly correlated to observed precipitation (see Appendix A).

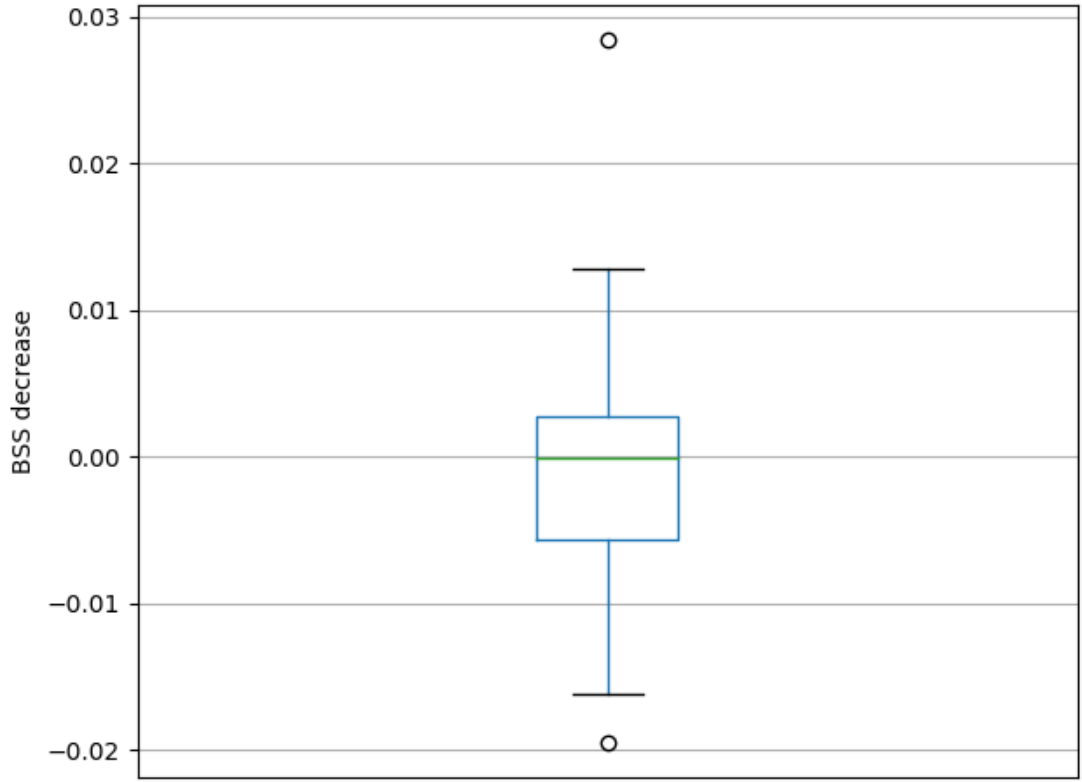

Figure 20: BSS decrease from removing the GEFS first quartile, GEFS third quartile, and the GEFS control forecast as features (random forests, all lead times, all thresholds)

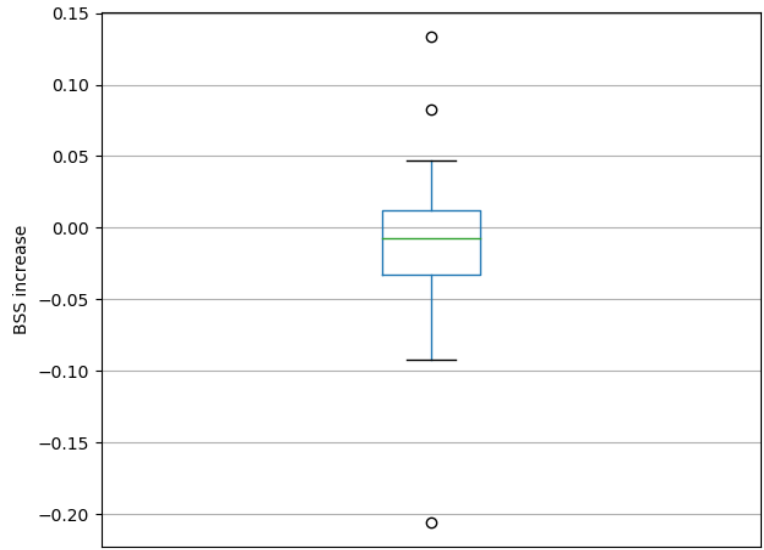

Figure 21: BSS increase from doubling the CNN filter count (all lead times, all thresholds)

Doubling the number of filters does not improve performance, suggesting that the CNN is not limited by representation depth (Figure 21). In none of the CNN runs, the validation loss ever significantly increased while training loss went down. All loss curves were similar to Figure 22, giving no indication of overfitting whatsoever.

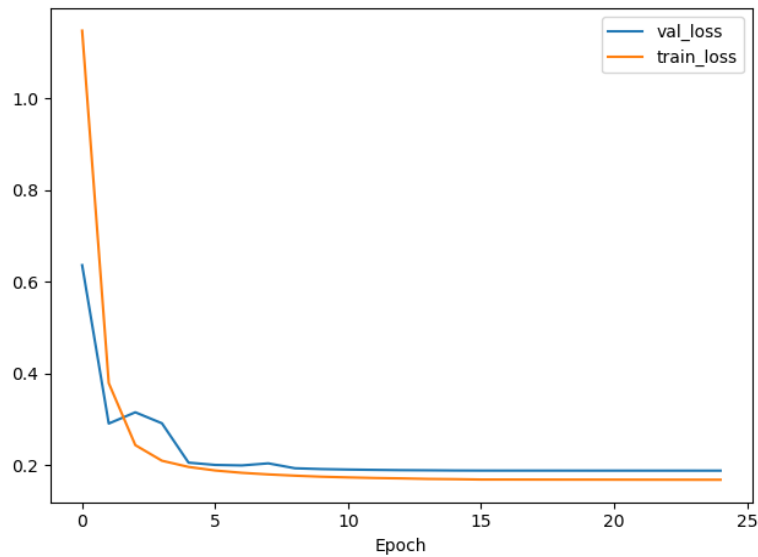

Figure 22: Training and validation loss for a single representative fold of the CNN

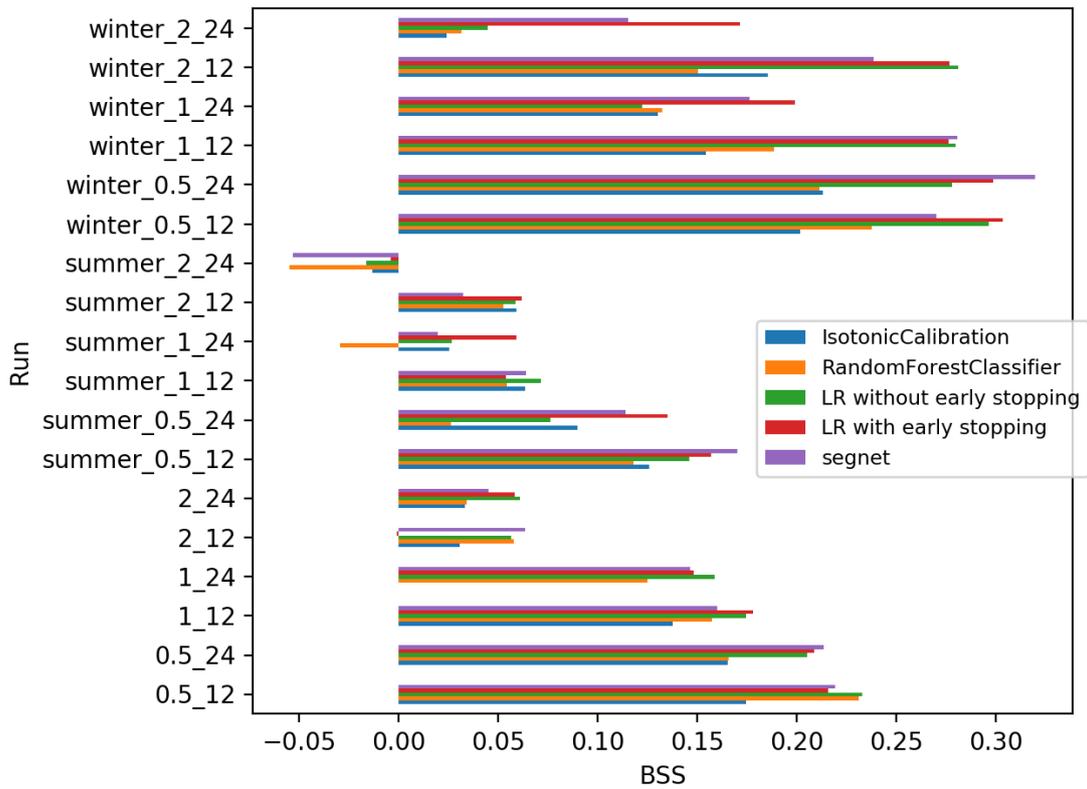

Figure 23: Test set performance for configurations with best validation score. `IsotonicCalibration` is isotonic regression fitted on either the GEFS ensemble mean or the HARMONIE forecast, whichever had better validation performance. Runs are formatted as `threshold_lead`, with an optional summer or winter prefix denoting the season-specific subexperiments.

Reported Brier skill scores vary wildly for the 1 and 2 mm subexperiments (Figure 23), presumably because of the small number of positive samples to evaluate on (Table 3). For the 0.5 mm threshold, the findings are more robust. Summer forecasts, forecasts for low thresholds, and 24-hour forecasts are less skillful than winter forecasts, high-threshold forecasts and 12-hour forecasts.

All post-processed forecasts generally outperformed the best-performing calibrated input forecast (`IsotonicCalibration` in Figure 23). None of the post-processing models performed much better than the others; logistic regression seemed most consistent.

# 5  Discussion

I set out to improve the post-processing of precipitation forecasts using CNNs. Instead of post-processing forecasts on a per-pixel basis, as is usually done when applying machine learning to meteorological post-processing, input forecast images were combined and transformed into output forecast images using fully convolutional neural networks.

Observations and forecasts were collected for the 2015-2017 period. Forecasts were combined and post-processed using logistic regression, random forests, and a CNN architecture based on SegNet. Hourly precipitation accumulations for Dutch land mass were forecasted 12 and 24 hours into the future. Output forecasts were probabilistic and binary—the task was to forecast whether precipitation would exceed a threshold, which was set to 0.5 mm, 1 mm or 2 mm. All post-processed forecasts generally outperformed calibrated input forecasts. None of the post-processing models performed much better than the others; logistic regression appeared to be most robust.

## 5.1  Measuring feature importance

Mean decrease in impurity (MDI) and permutation importances are gaining popularity as a measure of importance of individual features in meteorological model post-processing (e.g. Eccel et al. 2007; Straaten, Whan, and Schmeits 2018). Neither is a measure of the marginal benefit of a feature to forecasting performance. The marginal benefit of a feature is usually paramount in meteorological post-postprecessing, since many features are highly correlated to the ground truth but redundant.

Redundant features can be considered important by both MDI and permutation importances. While these metrics are able to assign low importances to features which are simply noisier versions of other features, they cannot solve the attribution problem for two equally informative, highly collinear features. This is clear in the trivial case where two features are duplicates. These features, being equally informative, are given the same score, even though one feature is strictly unnecessary.

A naive reading of the MDI scores would be that input forecasts are less important for the output forecast than non-forecast features. However, redundant features distort the feature importances. `xdim`, a feature roughly corresponding to latitude, was the feature with the highest MDI, higher than the feature importance of the HARMONIE forecast and the GEFS ensemble mean. This is presumably only so because `xdim` did not have to share its feature importance with highly correlated, redundant features, as was the case for the GEFS and HARMONIE forecasts (Appendix A). Ablation analyses seem quite rare in the field of meteorology, although feature selection schemes like stepwise regression are commonly used.

My recommendation is to perform an ablation analysis, whenever feasible, when making claims about the benefit of new meteorological features that are highly collinear to existing features. MDI and permutation importance are still useful for non-collinear features. If the model is wholly predictive rather than descriptive, redundant features can be eliminated using feature selection schemes such as stepwise regression or regularized random forests (Deng and Runger 2012).

## 5.2 Feature evaluation

The ablation analysis for logistic regression shows that there is no benefit to including GEFS ensemble statistics—the first quartile, third quartile, and the control forecast—when the ensemble mean is already included (Figure 17; Figure 20). Combining forecasts from different sources does increase performance. Logistic regression fit on both the HARMONIE forecast and the GEFS ensemble mean outperforms logistic regression fit on just one of the two. This is in line with other research on multi-model ensembles, which tend to outperform all individual members (Mylne, Evans, and Clark 2002; Weigel, Liniger, and Appenzeller 2008; Ziehmann 2000).

The high MDI for `xdim` is interesting because it suggests the random forest models an interaction between latitude and other variables; `xdim` is itself not correlated to the ground truth.

## 5.3 TPOT, random forests & logistic regression

TPOT, an automated machine learning suite, was used to test a large number of machine learning pipelines on a small subset of the data. The best-performing pipeline was a random forest configuration, which I went on to use in the rest of the thesis. In retrospect, performing a more extensive optimization run on just random forests may have been more fruitful.

Logistic regression performed at least as well as random forests in most subexperiments. In quite a few experiments, logistic regression performed much better. TPOT presumably did not find logistic regression as the optimal pipeline because of the fact that logistic regression requires unusually strong regularization ($100\,000 <= L2 <= 10\,000\,000$, depending on the subexperiment), which was not part of the TPOT search space (see Appendix C).

The requirement for strong regularization was discovered only after running TPOT. If a lower L2 was used for logistic regression, some derived input forecasts, which all ought to have positive coefficients, would be assigned negative coefficients.

The Keras implementation of logistic regression, using an SGD optimizer with early stopping, performed well with light regularization ($L2 = 0.01$), and may not have needed L2 regularization at all. After early stopping, the model would usually be restored to a single-digit epoch, whereas `sklearn` logistic regression was solved only after hundreds of epochs.

## 5.4 Calibration

All models turned out to be somewhat overconfident when forecasting high precipitation probabilities. This may be the case because the validation data the models were calibrated on was not independent from the training set. The data from every second day was chosen as the validation set, rather than using consecutive training and test sets. The reasoning for this is described in the method section.

In future work, it would be interesting to see how the two validation schemes compare in terms of calibration and Brier scores. Using one year for training and one year for validation may result in better calibration, but it might also result in worse performance, as the training and validation sets are less diverse.

## 5.5 CNN Performance

I expected CNNs to outperform logistic regression based on the results of CNNs in end-to-end short-term forecasting (Shi et al. 2017). However, over all subexperiments, per-pixel logistic regression and CNNs performed roughly the same. As such, many alterations were made to the SegNet-based CNN in mostly fruitless attempts to improve performance (Table 4).

Additionally, to verify that low performance was not the result of a mistake in the CNN pipeline or a result of an implementation detail, per-pixel logistic regression was implemented as a CNN with one 1x1 layer. This performed just as well as `sklearn` logistic regression using the per-pixel pipeline. Typically, logistic regression, which is a convex problem, is solved for the global minimum. Interestingly, early stopping in logistic regression fit using stochastic gradient descent appeared to make the need for L2 regularization much lower.

The CNN did not overfit on the training data at any point during development. Validation loss never went back up significantly during training. Nonetheless, the CNN did not seem constrained in its representational abilities. Doubling the filter depths did not increase performance, nor did increasing the number of layers. As a sanity check, the CNN was trained on 40 images. The CNN was able to overfit on this smaller set.

The CNN forecasts did not differ much visually from the per-pixel averaging performed by per-pixel logistic regression. No attempt at advanced spatial post-processing, such as correcting translation errors, was seen. This was partially the result of the decision to include a skip connection from the input to the last CNN layer. This did increase performance, but in retrospect, the failure of the CNN forecasts might have been more interpretable if the CNN was not able to rely on this connection during development. In the development tests without this skip connection, the output forecasts were just smeared-out, low-sharpness versions of the input forecasts.

It is likely that a weight assignment for the CNN exists which would give much better results. If that's the case, the model got stuck in a local minimum. The fact that CNNs generally do perform well on short-term end-to-end forecasting limits the possible explanations for this. For instance, the ground truth is based on noisy radar images. It is conceivable that the mislabeled samples resulting from noise in the ground truth make the loss landscape so jagged that the neural network gets stuck in a local minimum. However, if this were the case, CNNs for short-term forecasts would also be unable to learn anything, as they are typically also trained on radar observations. As far as I am aware, there has been no work training CNNs for short-term forecasts on the observation dataset I used, so the possibility exists that the disparity between my medium-term forecasts on the KNMI radar images and the successful short-term forecasts based on other datasets is due to differences in geography, climate, radar configuration, or radar calibration methods. To rule this out, it would be informative to have the work in this paper replicated on a dataset already successfully used in short-term forecasting.

A more likely reason I see for the disparity between short-term end-to-end systems and my post-processing CNN, is that the end-to-end systems—having no forecasts as input—are forced to develop a representation of how precipitation patterns develop, while the post-processing system is not. End-to-end systems output an entire run of forecasts, with one CNN responsible for forecasting all lead times. Each post-processing CNN only outputs forecasts for a single lead time, in line with how post-processing is usually performed in meteorology (e.g. Straaten, Whan, and Schmeits 2018). While this makes sense for random forests and logistic regression, separation by lead time does not allow the CNN to function as it does in end-to-end systems. The model has to learn how to post-process +12 hour precipitation forecasts without being trained how to post-process the intermediary forecasts. Because only the forecasts for the hours before and after the validation time were provided as an input, rather than all input forecasts up to the validation time, the CNN is unable to verify if the development of precipitation patterns is correct.

It might be illustrative to compare my model to a performant end-to-end nowcasting system. In a comparative benchmark study for nowcasting, a 2D CNN received five six-minute frames to forecast twenty six-minute frames (Shi et al. 2017). By comparison, my post-processing model for the +12-hour lead time did not have access to +1-hour to +9-hour forecasts, nor to the hidden infrastructure used to forecast previous lead times.

I included the observations for the hour before the initialization time and the forecasts for the hours before and after the validation time were included as features, but this is very different from the information an end-to-end system has available. I provided the observations for the hour before the initialization time as a single feature, while the end-to-end system gets multiple consecutive frames as input.

An alternative hypothesis is that the accumulation step of both input forecasts and ground truth may simply be too long (i.e. the frame rate may be too low) for the CNN to issue meaningful corrections. For a system like a CNN, which has no prior information, it might be easier to learn how clouds develop based on 5-minute accumulated observations than based on 1-hour accumulations.

Future work should look at using multiple 5-minute observations as input before initialization time, instead of including the precipitation in the hour before the initialization time as a single feature. Including all forecasts up to the validation time might also help. Additionally, a single CNN that post-processes an entire run of forecasts (i.e. the forecasts for all lead times) could be built, allowing the model to function as an end-to-end forecasting system whenever necessary. This model could be a 2D CNN, a 3D CNN, or any of the models benchmarked by Shi et al. (2017).

Additional data may also increase performance, as is often the case for neural networks. GEFS reforecasts and KNMI observations go back to 1986 and 2008, respectively, while KNMI rain reforecasts were only made for the 2015-2017 period. Future work can use the older GEFS forecasts and KNMI observations. One option is to only train on GEFS. Another is to pre-train the network on the older GEFS reforecasts, then expanding the input layer to take KNMI reforecasts as well.

Finally, to help the CNN form a more complete representation of atmospheric workings, additional meteorological variables could be included as input, such as pressure and wind speed.

# 6 Conclusions

I did not succeed in employing CNNs to improve the post-processing of precipitation forecasts compared to logistic regression. An architecture was built to combine and transform input forecasts from multiple sources into one output image. However, no evidence was found that CNNs perform better than per-pixel logistic regression for post-processing medium-term precipitation forecasts. All post-processing models, including CNNs, did generally achieve higher brier skill scores than the best calibrated input forecast.

Including the HARMONIE forecast in addition to the GEFS ensemble mean increased logistic regression performance by a large margin. This was expected because the regional HARMONIE model has a higher spatial and temporal resolution than GEFS, which is global. Interestingly, including the GEFS ensemble mean as an input to logistic regression resulted in improvement over a model fit on just the HARMONIE forecast. This is a testament to the strength of multi-model ensembles.

By contrast, including additional information about the GEFS ensemble, such as the first and third quartile, did not increase performance. A likely explanation for this is that all individual GEFS members are simulated using the same NWP model, limiting the ensemble spread. However, it could still be that features like the first and third quartile are useful for high-resolution ensemble forecasts, which were not used in this thesis.

Consistent with Herman and Schumacher (2018), L2 regularization improved brier skill scores for logistic regression by a large margin. Early stopping was found to work at least as well as L2 as a regularization measure when the logistic regression model was fit using stochastic gradient descent.

Random forests performed worse than logistic regression in most subexperiments. This may have been the result of limited hyperparameter optimization in the case of random forests. To evaluate the addition of highly-correlated, possibly redundant features in exploratory research, an ablation analysis was found to be more useful than MDI, the feature importance computed for random forests, a metric commonly reported in meteorological post-processing.

# 7 Attribution

Figure 4: en:User:Cburnett[7] / CC BY-SA

Figure 6: Liang Chen, Paul Bentley, Daniel Rueckert / CC BY 4.0

Figure 7: Mehrdad Yazdani[8] / CC BY-SA

---

[7]https://commons.wikimedia.org/wiki/File:Artificial_neural_network.svg
[8]https://commons.wikimedia.org/wiki/File:Example_architecture_of_U-Net_for_producing_k_256-by-256_image_masks_for_a_256-by-256_RGB_image.png

# Appendix A

|  | hm2 | hm1 | harmonie | hp1 | hp2 | ga_prev | gefs_avg | ga_next | gefs_control | gefs_q1 | gefs_q3 | gefs_t | init_obs | ydim | xdim | tdim | gefs_avg_lmax | ground truth |
|---|---|---|---|---|---|---|---|---|---|---|---|---|---|---|---|---|---|---|
| hm2 | 1.00 | 0.39 | 0.25 | 0.16 | 0.11 | 0.41 | 0.37 | 0.20 | 0.35 | 0.37 | 0.36 | 0.34 | 0.04 | 0.01 | -0.00 | 0.01 | 0.35 | 0.22 |
| hm1 | 0.39 | 1.00 | 0.37 | 0.23 | 0.16 | 0.32 | 0.43 | 0.25 | 0.42 | 0.43 | 0.42 | 0.39 | 0.05 | -0.01 | 0.00 | -0.00 | 0.39 | 0.29 |
| harmonie | 0.25 | 0.37 | 1.00 | 0.38 | 0.22 | 0.23 | 0.43 | 0.31 | 0.42 | 0.41 | 0.43 | 0.37 | 0.05 | -0.02 | 0.00 | -0.01 | 0.39 | 0.32 |
| hp1 | 0.16 | 0.23 | 0.38 | 1.00 | 0.38 | 0.17 | 0.40 | 0.37 | 0.39 | 0.37 | 0.40 | 0.34 | 0.03 | -0.02 | -0.00 | 0.01 | 0.38 | 0.28 |
| hp2 | 0.11 | 0.16 | 0.22 | 0.38 | 1.00 | 0.14 | 0.34 | 0.40 | 0.33 | 0.30 | 0.35 | 0.28 | 0.03 | -0.03 | -0.00 | 0.02 | 0.35 | 0.24 |
| ga_prev | 0.41 | 0.32 | 0.23 | 0.17 | 0.14 | 1.00 | 0.56 | 0.27 | 0.53 | 0.56 | 0.54 | 0.50 | 0.07 | 0.05 | -0.01 | -0.01 | 0.53 | 0.27 |
| gefs_avg | 0.37 | 0.43 | 0.43 | 0.40 | 0.34 | 0.56 | 1.00 | 0.64 | 0.98 | 0.97 | 0.99 | 0.87 | 0.08 | 0.02 | -0.00 | 0.01 | 0.85 | 0.47 |
| ga_next | 0.20 | 0.25 | 0.31 | 0.37 | 0.40 | 0.27 | 0.64 | 1.00 | 0.61 | 0.60 | 0.65 | 0.58 | 0.06 | 0.02 | -0.01 | 0.03 | 0.66 | 0.36 |
| gefs_control | 0.35 | 0.42 | 0.42 | 0.39 | 0.33 | 0.53 | 0.98 | 0.61 | 1.00 | 0.95 | 0.97 | 0.85 | 0.07 | 0.02 | -0.00 | 0.00 | 0.82 | 0.45 |
| gefs_q1 | 0.37 | 0.43 | 0.41 | 0.37 | 0.30 | 0.56 | 0.97 | 0.60 | 0.95 | 1.00 | 0.93 | 0.86 | 0.07 | 0.02 | -0.00 | 0.01 | 0.80 | 0.45 |
| gefs_q3 | 0.36 | 0.42 | 0.43 | 0.40 | 0.35 | 0.54 | 0.99 | 0.65 | 0.97 | 0.93 | 1.00 | 0.86 | 0.08 | 0.02 | -0.00 | 0.00 | 0.86 | 0.46 |
| gefs_t | 0.34 | 0.39 | 0.37 | 0.34 | 0.28 | 0.50 | 0.87 | 0.58 | 0.85 | 0.86 | 0.86 | 1.00 | 0.07 | 0.02 | -0.00 | 0.00 | 0.75 | 0.43 |
| init_obs | 0.04 | 0.05 | 0.05 | 0.03 | 0.03 | 0.07 | 0.08 | 0.06 | 0.07 | 0.07 | 0.08 | 0.07 | 1.00 | -0.02 | -0.00 | -0.00 | 0.09 | 0.05 |
| ydim | 0.01 | -0.01 | -0.02 | -0.02 | -0.03 | 0.05 | 0.02 | 0.02 | 0.02 | 0.02 | 0.02 | 0.02 | -0.02 | 1.00 | -0.00 | 0.00 | 0.04 | -0.01 |
| xdim | -0.00 | 0.00 | 0.00 | -0.00 | -0.00 | -0.01 | -0.00 | -0.01 | -0.00 | -0.00 | -0.00 | -0.00 | -0.00 | -0.00 | 1.00 | -0.36 | -0.04 | -0.00 |
| tdim | 0.01 | -0.00 | -0.01 | 0.01 | 0.02 | -0.01 | 0.01 | 0.03 | 0.00 | 0.01 | 0.00 | 0.00 | -0.00 | 0.00 | -0.36 | 1.00 | 0.00 | -0.00 |
| gefs_avg_lmax | 0.35 | 0.39 | 0.39 | 0.38 | 0.35 | 0.53 | 0.85 | 0.66 | 0.82 | 0.80 | 0.86 | 0.75 | 0.09 | 0.04 | -0.04 | 0.00 | 1.00 | 0.43 |
| ground truth | 0.22 | 0.29 | 0.32 | 0.28 | 0.24 | 0.27 | 0.47 | 0.36 | 0.45 | 0.45 | 0.46 | 0.43 | 0.05 | -0.01 | -0.00 | -0.00 | 0.43 | 1.00 |

Pearson's correlation matrix for l = +12. gefs_t is computed using a 0.5 mm threshold.

# Appendix B: Best-performing configurations

Table 6: Many configurations were tried. `pretty_config` is the model configuration which had the best `valmedian`, i.e. the best median brier skill score on the validation set. `testmedian` is the corresponding median brier skill score of the calibrated model on the test set.

| season | lead | threshold | pretty_config | valmedian | testmedian |
|---|---|---|---|---|---|
| all | 12 | 0.5 | IsotonicCalibration() / ['area_harmonie'] | 0.2061 | 0.1747 |
| | | | RandomForestClassifier(n_estimators=100) / all features | 0.2365 | 0.2312 |
| | | | linear_model.LogisticRegression(C=.00001) / all features | 0.2284 | 0.2332 |
| | | | log_reg('REWEIGHT': False, 'SELECTED': 'nn_selected') / all features | 0.1912 | 0.2158 |
| | | | segnet('FILTER_MUL': 1, 'KERNEL': 3, 'ODD_POOL': True) / all features | 0.2137 | 0.2196 |
| | | 1 | IsotonicCalibration() / ['area_harmonie'] | 0.1664 | 0.1378 |
| | | | RandomForestClassifier(n_estimators=100) / all features | 0.1091 | 0.1574 |
| | | | linear_model.LogisticRegression(C=.00001) / ['area_harmonie', 'area_gefs_avg', 'area_gefs_avg_lmax'] | 0.1770 | 0.1745 |
| | | | log_reg('REWEIGHT': False, 'SELECTED': 'nn_selected') / all features | 0.1904 | 0.1781 |
| | | | segnet('FILTER_MUL': 1, 'KERNEL': 3, 'ODD_POOL': True) / all features | 0.1775 | 0.1600 |
| | | 2 | IsotonicCalibration() / ['area_harmonie'] | 0.0203 | 0.0305 |
| | | | RandomForestClassifier(n_estimators=100) / all features | -0.0373 | 0.0578 |
| | | | linear_model.LogisticRegression(C=.00001) / ['area_harmonie', 'area_gefs_avg', 'area_ga_prev'] | 0.0394 | 0.0565 |
| | | | log_reg('REWEIGHT': False, 'SELECTED': 'nn_selected') / all features | -0.0176 | -0.0009 |
| | | | segnet('FILTER_MUL': 2, 'KERNEL': 3, 'ODD_POOL': True) / all features | 0.1043 | 0.0635 |
| | 24 | 0.5 | IsotonicCalibration() / ['area_harmonie'] | 0.1750 | 0.1654 |
| | | | RandomForestClassifier(n_estimators=100) / all features | 0.1862 | 0.1660 |
| | | | linear_model.LogisticRegression(C=.001) / ['area_harmonie', 'area_gefs_avg', 'area_gefs_control'] | 0.1991 | 0.2052 |
| | | | log_reg('REWEIGHT': False, 'SELECTED': 'nn_selected') / all features | 0.2034 | 0.2087 |
| | | | segnet('FILTER_MUL': 2, 'KERNEL': 3, 'ODD_POOL': True) / all features | 0.2422 | 0.2136 |
| | | 1 | RandomForestClassifier(n_estimators=100) / all features | 0.1117 | 0.1252 |
| | | | linear_model.LogisticRegression(C=.00001) / ['area_harmonie', 'area_gefs_avg', 'area_hp1'] | 0.1262 | 0.1590 |
| | | | log_reg('REWEIGHT': False, 'SELECTED': 'nn_selected') / all features | 0.1292 | 0.1484 |
| | | | segnet('FILTER_MUL': 2, 'KERNEL': 3, 'ODD_POOL': False) / all features | 0.1500 | 0.1467 |
| | | 2 | IsotonicCalibration() / ['area_harmonie'] | 0.0181 | 0.0335 |
| | | | RandomForestClassifier(n_estimators=100) / all features | -0.0105 | 0.0343 |
| | | | linear_model.LogisticRegression(C=.00001) / ['area_harmonie', 'area_gefs_avg', 'area_hm1'] | 0.0455 | 0.0608 |
| | | | log_reg('REWEIGHT': False, 'SELECTED': 'nn_selected') / all features | 0.0196 | 0.0582 |
| | | | segnet('FILTER_MUL': 2, 'KERNEL': 3, 'ODD_POOL': True) / all features | 0.0047 | 0.0450 |
| summer | 12 | 0.5 | IsotonicCalibration() / ['area_harmonie'] | 0.0822 | 0.1261 |
| | | | RandomForestClassifier(n_estimators=100) / all features | 0.1191 | 0.1180 |
| | | | linear_model.LogisticRegression(C=.001) / ['area_harmonie', 'area_gefs_avg', 'area_ga_prev'] | 0.0918 | 0.1462 |
| | | | log_reg('REWEIGHT': False, 'SELECTED': 'nn_selected') / all features | 0.0960 | 0.1569 |
| | | | segnet('FILTER_MUL': 2, 'KERNEL': 3, 'ODD_POOL': False) / all features | 0.1404 | 0.1700 |
| | | 1 | IsotonicCalibration() / ['area_harmonie'] | 0.0199 | 0.0638 |
| | | | RandomForestClassifier(n_estimators=100) / all features | 0.0213 | 0.0545 |
| | | | linear_model.LogisticRegression(C=.001) / ['area_harmonie', 'area_gefs_avg', 'area_ga_prev'] | 0.0317 | 0.0717 |
| | | | log_reg('REWEIGHT': False, 'SELECTED': 'nn_selected') / all features | 0.0361 | 0.0540 |
| | | | segnet('FILTER_MUL': 1, 'KERNEL': 3, 'ODD_POOL': True) / all features | 0.0114 | 0.0639 |
| | | 2 | IsotonicCalibration() / ['area_harmonie'] | 0.0316 | 0.0592 |
| | | | RandomForestClassifier(n_estimators=100) / all features | 0.0040 | 0.0527 |
| | | | linear_model.LogisticRegression(C=.00001) / all features | 0.0689 | 0.0587 |
| | | | log_reg('REWEIGHT': False, 'SELECTED': 'nn_selected') / all features | 0.0361 | 0.0617 |
| | | | segnet('FILTER_MUL': 1, 'KERNEL': 3, 'ODD_POOL': True) / all features | -0.0136 | 0.0323 |
| | 24 | 0.5 | IsotonicCalibration() / ['area_harmonie'] | 0.0254 | 0.0901 |
| | | | RandomForestClassifier(n_estimators=100) / all features | 0.0067 | 0.0265 |
| | | | linear_model.LogisticRegression(C=.00001) / all features | 0.0468 | 0.0764 |
| | | | log_reg('REWEIGHT': False, 'SELECTED': 'nn_selected') / all features | 0.0417 | 0.1350 |
| | | | segnet('FILTER_MUL': 1, 'KERNEL': 3, 'ODD_POOL': True) / all features | 0.0324 | 0.1141 |
| | | 1 | IsotonicCalibration() / ['area_harmonie'] | -0.0124 | 0.0253 |
| | | | RandomForestClassifier(n_estimators=100) / all features | -0.1377 | -0.0295 |
| | | | linear_model.LogisticRegression(C=.00001) / ['area_harmonie', 'area_gefs_avg', 'area_ga_next'] | 0.0550 | 0.0266 |
| | | | log_reg('REWEIGHT': False, 'SELECTED': 'nn_selected') / all features | 0.0403 | 0.0594 |
| | | | segnet('FILTER_MUL': 1, 'KERNEL': 3, 'ODD_POOL': True) / all features | -0.0535 | 0.0197 |
| | | 2 | IsotonicCalibration() / all features | -0.0207 | -0.0131 |
| | | | RandomForestClassifier(n_estimators=100) / ['area_harmonie', 'area_gefs_avg', 'xdim', 'ydim'] | -0.2457 | -0.0547 |
| | | | linear_model.LogisticRegression(C=.00001) / ['area_harmonie', 'area_gefs_avg', 'area_ga_next'] | 0.0009 | -0.0164 |
| | | | log_reg('REWEIGHT': False, 'SELECTED': 'nn_selected') / all features | 0.0182 | -0.0039 |
| | | | segnet('FILTER_MUL': 2, 'KERNEL': 3, 'ODD_POOL': False) / all features | 0.0132 | -0.0532 |
| winter | 12 | 0.5 | IsotonicCalibration() / ['area_harmonie'] | 0.1098 | 0.2020 |
| | | | RandomForestClassifier(n_estimators=100) / all features | 0.1753 | 0.2377 |
| | | | linear_model.LogisticRegression(C=.001) / ['area_harmonie', 'area_gefs_avg', 'area_ga_prev'] | 0.1955 | 0.2965 |
| | | | log_reg('REWEIGHT': False, 'SELECTED': 'nn_selected') / all features | 0.2360 | 0.3038 |
| | | | segnet('FILTER_MUL': 1, 'KERNEL': 3, 'ODD_POOL': False) / all features | 0.2826 | 0.2701 |
| | | 1 | IsotonicCalibration() / ['area_harmonie'] | -0.0494 | 0.1543 |
| | | | RandomForestClassifier(n_estimators=100) / all features | 0.0895 | 0.1887 |
| | | | linear_model.LogisticRegression(C=.001) / ['area_harmonie', 'area_gefs_avg', 'area_ga_prev'] | 0.2427 | 0.2799 |
| | | | log_reg('REWEIGHT': False, 'SELECTED': 'nn_selected') / all features | 0.2096 | 0.2763 |
| | | | segnet('FILTER_MUL': 2, 'KERNEL': 3, 'ODD_POOL': False) / all features | 0.2628 | 0.2810 |
| | | 2 | IsotonicCalibration() / ['area_harmonie', 'area_gefs_avg', 'area_hp2'] | -0.1068 | 0.1856 |
| | | | RandomForestClassifier(n_estimators=100) / all features | -0.0294 | 0.1505 |
| | | | linear_model.LogisticRegression(C=.001) / ['area_harmonie', 'area_gefs_avg', 'area_gefs_control'] | 0.0787 | 0.2812 |
| | | | log_reg('REWEIGHT': False, 'SELECTED': 'nn_selected') / all features | -0.1007 | 0.2769 |
| | | | segnet('FILTER_MUL': 2, 'KERNEL': 3, 'ODD_POOL': True) / all features | 0.1560 | 0.2386 |
| | 24 | 0.5 | IsotonicCalibration() / ['area_harmonie'] | 0.2941 | 0.2134 |
| | | | RandomForestClassifier(n_estimators=100) / all features | 0.3532 | 0.2116 |
| | | | linear_model.LogisticRegression(C=.001) / ['area_harmonie', 'area_gefs_avg', 'area_ga_prev'] | 0.3826 | 0.2780 |
| | | | log_reg('REWEIGHT': False, 'SELECTED': 'nn_selected') / all features | 0.4054 | 0.2987 |
| | | | segnet('FILTER_MUL': 2, 'KERNEL': 3, 'ODD_POOL': True) / all features | 0.4228 | 0.3200 |
| | | 1 | IsotonicCalibration() / ['area_harmonie'] | 0.2645 | 0.1304 |
| | | | RandomForestClassifier(n_estimators=100) / all features | 0.3051 | 0.1323 |
| | | | linear_model.LogisticRegression(C=.1) / ['area_harmonie', 'area_gefs_avg', 'area_ga_prev'] | 0.3420 | 0.1222 |
| | | | log_reg('REWEIGHT': False, 'SELECTED': 'nn_selected') / all features | 0.3800 | 0.1990 |
| | | | segnet('FILTER_MUL': 2, 'KERNEL': 3, 'ODD_POOL': True) / all features | 0.3837 | 0.1762 |
| | | 2 | IsotonicCalibration() / ['area_harmonie', 'area_gefs_avg', 'area_hp2'] | 0.1586 | 0.0239 |
| | | | RandomForestClassifier(n_estimators=100) / all features | 0.1638 | 0.0316 |
| | | | linear_model.LogisticRegression(C=.001) / ['area_harmonie', 'area_gefs_avg', 'area_gefs_avg_lmax'] | 0.1582 | 0.0447 |
| | | | log_reg('REWEIGHT': False, 'SELECTED': 'nn_selected') / all features | 0.3343 | 0.1715 |
| | | | segnet('FILTER_MUL': 2, 'KERNEL': 3, 'ODD_POOL': True) / all features | 0.3343 | 0.1153 |

# Appendix C: Hyperparameter grid for TPOT

```
{
    'sklearn.naive_bayes.GaussianNB': {
    },

    'sklearn.naive_bayes.BernoulliNB': {
        'alpha': [1e-3, 1e-2, 1e-1, 1., 10., 100.],
        'fit_prior': [True, False]
    },

    'sklearn.naive_bayes.MultinomialNB': {
        'alpha': [1e-3, 1e-2, 1e-1, 1., 10., 100.],
        'fit_prior': [True, False]
    },

    'sklearn.tree.DecisionTreeClassifier': {
        'criterion': ["gini", "entropy"],
        'max_depth': range(1, 11),
        'min_samples_split': range(2, 21),
        'min_samples_leaf': range(1, 21)
    },

    'sklearn.neighbors.KNeighborsClassifier': {
        'n_neighbors': range(1, 101),
        'weights': ["uniform", "distance"],
        'p': [1, 2]
    },

    'sklearn.linear_model.LogisticRegression': {
        'penalty': ["l1", "l2"],
        'C': [1e-4, 1e-3, 1e-2, 1e-1, 0.5, 1., 5., 10., 15., 20., 25.],
        'dual': [True, False]
    },

    # Preprocesssors
    'sklearn.preprocessing.Binarizer': {
        'threshold': np.arange(0.0, 1.01, 0.05)
    },

    'sklearn.cluster.FeatureAgglomeration': {
        'linkage': ['ward', 'complete', 'average'],
        'affinity': ['euclidean', 'l1', 'l2', 'manhattan', 'cosine']
    },

    'sklearn.preprocessing.MaxAbsScaler': {
    },

    'sklearn.preprocessing.MinMaxScaler': {
    },

    'sklearn.preprocessing.Normalizer': {
        'norm': ['l1', 'l2', 'max']
    },

    'sklearn.decomposition.PCA': {
        'svd_solver': ['randomized'],
        'iterated_power': range(1, 11)
    },

    'sklearn.kernel_approximation.RBFSampler': {
        'gamma': np.arange(0.0, 1.01, 0.05)
    },

    'sklearn.preprocessing.RobustScaler': {
    },

    'sklearn.preprocessing.StandardScaler': {
    },

    'tpot.builtins.ZeroCount': {
    },

    # Selectors
    'sklearn.feature_selection.SelectFwe': {
        'alpha': np.arange(0, 0.05, 0.001),
        'score_func': {
            'sklearn.feature_selection.f_classif': None
        }
    },

    'sklearn.feature_selection.SelectPercentile': {
        'percentile': range(1, 100),
        'score_func': {
            'sklearn.feature_selection.f_classif': None
        }
    },

    'sklearn.feature_selection.VarianceThreshold': {
        'threshold': [0.0001, 0.0005, 0.001, 0.005, 0.01, 0.05, 0.1, 0.2]
    },

    'xgboost.XGBClassifier': {
        'n_estimators': [100],
```

```
            'max_depth': range(1, 11),
            'learning_rate': [1e-3, 1e-2, 1e-1, 0.5, 1.],
            'subsample': np.arange(0.05, 1.01, 0.05),
            'min_child_weight': range(1, 21),
            'nthread': [1]
        },
    'sklearn.ensemble.RandomForestClassifier': {
            'n_estimators': [100],
            'criterion': ["gini", "entropy"],
            'max_features': np.arange(0.05, 1.01, 0.05),
            'min_samples_split':  range(2, 21),
            'min_samples_leaf':  range(1, 21),
            'bootstrap': [True, False]
        }
}
```

# Appendix D: Final SegNet-based model

```
__________________________________________________________________________________________________
Layer (type)                    Output Shape         Param #     Connected to
==================================================================================================
input_7 (InputLayer)            [(5, 384, 384, 14)]  0
__________________________________________________________________________________________________
conv2d_56 (Conv2D)              (5, 384, 384, 8)     1016        input_7[0][0]
__________________________________________________________________________________________________
batch_normalization_56 (BatchNo (5, 384, 384, 8)     32          conv2d_56[0][0]
__________________________________________________________________________________________________
activation_56 (Activation)      (5, 384, 384, 8)     0           batch_normalization_56[0][0]
__________________________________________________________________________________________________
max_pooling_with_argmax2d_14 (M [(5, 192, 192, 8), ( 0           activation_56[0][0]
__________________________________________________________________________________________________
conv2d_57 (Conv2D)              (5, 192, 192, 8)     584         max_pooling_with_argmax2d_14[0][0
__________________________________________________________________________________________________
batch_normalization_57 (BatchNo (5, 192, 192, 8)     32          conv2d_57[0][0]
__________________________________________________________________________________________________
activation_57 (Activation)      (5, 192, 192, 8)     0           batch_normalization_57[0][0]
__________________________________________________________________________________________________
max_pooling_with_argmax2d_15 (M [(5, 96, 96, 8), (5, 0           activation_57[0][0]
__________________________________________________________________________________________________
conv2d_58 (Conv2D)              (5, 96, 96, 16)      1168        max_pooling_with_argmax2d_15[0][0
__________________________________________________________________________________________________
batch_normalization_58 (BatchNo (5, 96, 96, 16)      64          conv2d_58[0][0]
__________________________________________________________________________________________________
activation_58 (Activation)      (5, 96, 96, 16)      0           batch_normalization_58[0][0]
__________________________________________________________________________________________________
conv2d_59 (Conv2D)              (5, 96, 96, 16)      2320        activation_58[0][0]
__________________________________________________________________________________________________
batch_normalization_59 (BatchNo (5, 96, 96, 16)      64          conv2d_59[0][0]
__________________________________________________________________________________________________
activation_59 (Activation)      (5, 96, 96, 16)      0           batch_normalization_59[0][0]
__________________________________________________________________________________________________
max_pooling_with_argmax2d_16 (M [(5, 48, 48, 16), (5 0           activation_59[0][0]
__________________________________________________________________________________________________
conv2d_60 (Conv2D)              (5, 48, 48, 32)      4640        max_pooling_with_argmax2d_16[0][0
__________________________________________________________________________________________________
batch_normalization_60 (BatchNo (5, 48, 48, 32)      128         conv2d_60[0][0]
__________________________________________________________________________________________________
activation_60 (Activation)      (5, 48, 48, 32)      0           batch_normalization_60[0][0]
__________________________________________________________________________________________________
max_pooling_with_argmax2d_17 (M [(5, 24, 24, 32), (5 0           activation_60[0][0]
__________________________________________________________________________________________________
conv2d_61 (Conv2D)              (5, 24, 24, 32)      9248        max_pooling_with_argmax2d_17[0][0
__________________________________________________________________________________________________
batch_normalization_61 (BatchNo (5, 24, 24, 32)      128         conv2d_61[0][0]
__________________________________________________________________________________________________
activation_61 (Activation)      (5, 24, 24, 32)      0           batch_normalization_61[0][0]
__________________________________________________________________________________________________
conv2d_62 (Conv2D)              (5, 24, 24, 32)      9248        activation_61[0][0]
__________________________________________________________________________________________________
batch_normalization_62 (BatchNo (5, 24, 24, 32)      128         conv2d_62[0][0]
__________________________________________________________________________________________________
activation_62 (Activation)      (5, 24, 24, 32)      0           batch_normalization_62[0][0]
__________________________________________________________________________________________________
max_pooling_with_argmax2d_18 (M [(5, 12, 12, 32), (5 0           activation_62[0][0]
__________________________________________________________________________________________________
conv2d_63 (Conv2D)              (5, 12, 12, 64)      18496       max_pooling_with_argmax2d_18[0][0
__________________________________________________________________________________________________
batch_normalization_63 (BatchNo (5, 12, 12, 64)      256         conv2d_63[0][0]
__________________________________________________________________________________________________
activation_63 (Activation)      (5, 12, 12, 64)      0           batch_normalization_63[0][0]
__________________________________________________________________________________________________
conv2d_64 (Conv2D)              (5, 12, 12, 64)      36928       activation_63[0][0]
__________________________________________________________________________________________________
batch_normalization_64 (BatchNo (5, 12, 12, 64)      256         conv2d_64[0][0]
__________________________________________________________________________________________________
activation_64 (Activation)      (5, 12, 12, 64)      0           batch_normalization_64[0][0]
__________________________________________________________________________________________________
conv2d_65 (Conv2D)              (5, 12, 12, 64)      36928       activation_64[0][0]
__________________________________________________________________________________________________
batch_normalization_65 (BatchNo (5, 12, 12, 64)      256         conv2d_65[0][0]
__________________________________________________________________________________________________
```

```
activation_65 (Activation)      (5, 12, 12, 64)     0       batch_normalization_65[0][0]
________________________________________________________________________________________
max_pooling_with_argmax2d_19 (M [(5, 6, 6, 64), (5, 0       activation_65[0][0]
________________________________________________________________________________________
conv2d_66 (Conv2D)              (5, 6, 6, 64)       36928   max_pooling_with_argmax2d_19[0][0
________________________________________________________________________________________
batch_normalization_66 (BatchNo (5, 6, 6, 64)       256     conv2d_66[0][0]
________________________________________________________________________________________
activation_66 (Activation)      (5, 6, 6, 64)       0       batch_normalization_66[0][0]
________________________________________________________________________________________
conv2d_67 (Conv2D)              (5, 6, 6, 64)       36928   activation_66[0][0]
________________________________________________________________________________________
batch_normalization_67 (BatchNo (5, 6, 6, 64)       256     conv2d_67[0][0]
________________________________________________________________________________________
activation_67 (Activation)      (5, 6, 6, 64)       0       batch_normalization_67[0][0]
________________________________________________________________________________________
conv2d_68 (Conv2D)              (5, 6, 6, 64)       36928   activation_67[0][0]
________________________________________________________________________________________
batch_normalization_68 (BatchNo (5, 6, 6, 64)       256     conv2d_68[0][0]
________________________________________________________________________________________
activation_68 (Activation)      (5, 6, 6, 64)       0       batch_normalization_68[0][0]
________________________________________________________________________________________
max_pooling_with_argmax2d_20 (M [(5, 3, 3, 64), (5, 0       activation_68[0][0]
________________________________________________________________________________________
conv2d_69 (Conv2D)              (5, 3, 3, 64)       36928   max_pooling_with_argmax2d_20[0][0
________________________________________________________________________________________
batch_normalization_69 (BatchNo (5, 3, 3, 64)       256     conv2d_69[0][0]
________________________________________________________________________________________
activation_69 (Activation)      (5, 3, 3, 64)       0       batch_normalization_69[0][0]
________________________________________________________________________________________
max_unpooling2d_14 (MaxUnpoolin (5, 6, 6, 64)       0       activation_69[0][0]
                                                            max_pooling_with_argmax2d_20[0][1
                                                            activation_68[0][0]
________________________________________________________________________________________
conv2d_70 (Conv2D)              (5, 6, 6, 64)       36928   max_unpooling2d_14[0][0]
________________________________________________________________________________________
batch_normalization_70 (BatchNo (5, 6, 6, 64)       256     conv2d_70[0][0]
________________________________________________________________________________________
activation_70 (Activation)      (5, 6, 6, 64)       0       batch_normalization_70[0][0]
________________________________________________________________________________________
conv2d_71 (Conv2D)              (5, 6, 6, 64)       36928   activation_70[0][0]
________________________________________________________________________________________
batch_normalization_71 (BatchNo (5, 6, 6, 64)       256     conv2d_71[0][0]
________________________________________________________________________________________
activation_71 (Activation)      (5, 6, 6, 64)       0       batch_normalization_71[0][0]
________________________________________________________________________________________
conv2d_72 (Conv2D)              (5, 6, 6, 64)       36928   activation_71[0][0]
________________________________________________________________________________________
batch_normalization_72 (BatchNo (5, 6, 6, 64)       256     conv2d_72[0][0]
________________________________________________________________________________________
activation_72 (Activation)      (5, 6, 6, 64)       0       batch_normalization_72[0][0]
________________________________________________________________________________________
max_unpooling2d_15 (MaxUnpoolin (5, 12, 12, 64)     0       activation_72[0][0]
                                                            max_pooling_with_argmax2d_19[0][1
                                                            activation_65[0][0]
________________________________________________________________________________________
conv2d_73 (Conv2D)              (5, 12, 12, 64)     36928   max_unpooling2d_15[0][0]
________________________________________________________________________________________
batch_normalization_73 (BatchNo (5, 12, 12, 64)     256     conv2d_73[0][0]
________________________________________________________________________________________
activation_73 (Activation)      (5, 12, 12, 64)     0       batch_normalization_73[0][0]
________________________________________________________________________________________
conv2d_74 (Conv2D)              (5, 12, 12, 64)     36928   activation_73[0][0]
________________________________________________________________________________________
batch_normalization_74 (BatchNo (5, 12, 12, 64)     256     conv2d_74[0][0]
________________________________________________________________________________________
activation_74 (Activation)      (5, 12, 12, 64)     0       batch_normalization_74[0][0]
________________________________________________________________________________________
conv2d_75 (Conv2D)              (5, 12, 12, 32)     18464   activation_74[0][0]
________________________________________________________________________________________
batch_normalization_75 (BatchNo (5, 12, 12, 32)     128     conv2d_75[0][0]
________________________________________________________________________________________
activation_75 (Activation)      (5, 12, 12, 32)     0       batch_normalization_75[0][0]
________________________________________________________________________________________
max_unpooling2d_16 (MaxUnpoolin (5, 24, 24, 32)     0       activation_75[0][0]
                                                            max_pooling_with_argmax2d_18[0][1
                                                            activation_62[0][0]
________________________________________________________________________________________
conv2d_76 (Conv2D)              (5, 24, 24, 32)     9248    max_unpooling2d_16[0][0]
________________________________________________________________________________________
batch_normalization_76 (BatchNo (5, 24, 24, 32)     128     conv2d_76[0][0]
________________________________________________________________________________________
activation_76 (Activation)      (5, 24, 24, 32)     0       batch_normalization_76[0][0]
________________________________________________________________________________________
max_unpooling2d_17 (MaxUnpoolin (5, 48, 48, 32)     0       activation_76[0][0]
                                                            max_pooling_with_argmax2d_17[0][1
                                                            activation_60[0][0]
________________________________________________________________________________________
conv2d_77 (Conv2D)              (5, 48, 48, 32)     9248    max_unpooling2d_17[0][0]
________________________________________________________________________________________
batch_normalization_77 (BatchNo (5, 48, 48, 32)     128     conv2d_77[0][0]
________________________________________________________________________________________
activation_77 (Activation)      (5, 48, 48, 32)     0       batch_normalization_77[0][0]
________________________________________________________________________________________
conv2d_78 (Conv2D)              (5, 48, 48, 16)     4624    activation_77[0][0]
________________________________________________________________________________________
batch_normalization_78 (BatchNo (5, 48, 48, 16)     64      conv2d_78[0][0]
```

```
----------------------------------------------------------------------------------------------
activation_78 (Activation)     (5, 48, 48, 16)      0         batch_normalization_78[0][0]
----------------------------------------------------------------------------------------------
max_unpooling2d_18 (MaxUnpoolin (5, 96, 96, 16)     0         activation_78[0][0]
                                                              max_pooling_with_argmax2d_16[0][1
                                                              activation_59[0][0]
----------------------------------------------------------------------------------------------
conv2d_79 (Conv2D)             (5, 96, 96, 16)      2320      max_unpooling2d_18[0][0]
----------------------------------------------------------------------------------------------
batch_normalization_79 (BatchNo (5, 96, 96, 16)     64        conv2d_79[0][0]
----------------------------------------------------------------------------------------------
activation_79 (Activation)     (5, 96, 96, 16)      0         batch_normalization_79[0][0]
----------------------------------------------------------------------------------------------
conv2d_80 (Conv2D)             (5, 96, 96, 8)       1160      activation_79[0][0]
----------------------------------------------------------------------------------------------
batch_normalization_80 (BatchNo (5, 96, 96, 8)      32        conv2d_80[0][0]
----------------------------------------------------------------------------------------------
activation_80 (Activation)     (5, 96, 96, 8)       0         batch_normalization_80[0][0]
----------------------------------------------------------------------------------------------
max_unpooling2d_19 (MaxUnpoolin (5, 192, 192, 8)    0         activation_80[0][0]
                                                              max_pooling_with_argmax2d_15[0][1
                                                              activation_57[0][0]
----------------------------------------------------------------------------------------------
conv2d_81 (Conv2D)             (5, 192, 192, 8)     584       max_unpooling2d_19[0][0]
----------------------------------------------------------------------------------------------
batch_normalization_81 (BatchNo (5, 192, 192, 8)    32        conv2d_81[0][0]
----------------------------------------------------------------------------------------------
activation_81 (Activation)     (5, 192, 192, 8)     0         batch_normalization_81[0][0]
----------------------------------------------------------------------------------------------
max_unpooling2d_20 (MaxUnpoolin (5, 384, 384, 8)    0         activation_81[0][0]
                                                              max_pooling_with_argmax2d_14[0][1
                                                              activation_56[0][0]
----------------------------------------------------------------------------------------------
conv2d_82 (Conv2D)             (5, 384, 384, 8)     584       max_unpooling2d_20[0][0]
----------------------------------------------------------------------------------------------
batch_normalization_82 (BatchNo (5, 384, 384, 8)    32        conv2d_82[0][0]
----------------------------------------------------------------------------------------------
activation_82 (Activation)     (5, 384, 384, 8)     0         batch_normalization_82[0][0]
----------------------------------------------------------------------------------------------
input_9 (InputLayer)           [(5, 384, 384, 3)]   0
----------------------------------------------------------------------------------------------
concatenate_2 (Concatenate)    (5, 384, 384, 25)    0         activation_82[0][0]
                                                              input_7[0][0]
                                                              input_9[0][0]
----------------------------------------------------------------------------------------------
conv2d_83 (Conv2D)             (5, 384, 384, 1)     26        concatenate_2[0][0]
----------------------------------------------------------------------------------------------
batch_normalization_83 (BatchNo (5, 384, 384, 1)    4         conv2d_83[0][0]
----------------------------------------------------------------------------------------------
reshape_2 (Reshape)            (5, 147456)          0         batch_normalization_83[0][0]
----------------------------------------------------------------------------------------------
activation_83 (Activation)     (5, 147456)          0         reshape_2[0][0]
----------------------------------------------------------------------------------------------
input_8 (InputLayer)           [(5, 147456)]        0
----------------------------------------------------------------------------------------------
lambda_2 (Lambda)              (5, 38020)           0         activation_83[0][0]
==============================================================================================
Total params: 503,446
Trainable params: 501,316
Non-trainable params: 2,130
----------------------------------------------------------------------------------------------
```

# Appendix E: Details for Reproduction

Reprojecting to the unusual observation grid was unexpectedly challenging, taking almost a month. Many reprojection tools, such as `cdo`, did not seem to be able to handle the observation grid's unusual polar stereographic projection. The extent of the observation grid was defined as such that the latitude decreases as the y-coordinate on the projected grid increases, resulting in a y-flipped image. Finally, a working program using the `gdal` suite was constructed.

Temporarily aligning forecasts and observations also proved to be difficult. The date format used for the hourly accumulated observations is uncommon, as midnight is denoted as 24:00 of the day before rather than 00:00 of the day after. Beware that date parsers interpret this incorrectly. There appears to be no standard as to whether dates in file names denote the start of an accumulated period or the end.

The forecasts for higher and lower lead times were used to verify the correct alignment of forecasts and observations. The correlations between the +12h observations and the +11h, +12h, and

+13h HARMONIE forecasts were computed, as well as for the next, current and previous GEFS blocks. As expected, the +12h forecasts had the strongest correlation with the +12h observations (see Appendix A).

Both sources of input forecasts used in this thesis store the accumulated predicted precipitation rather than the hourly predicted precipitation. This accumulation is reset to 0 every $n$ hours, where $n$ is the accumulation window shown in Table 1.

Hourly precipitation forecasts were computed from the forecasted precipitation accumulations. In the case of the HARMONIE forecasts, which have an hourly temporal resolution and an accumulation window equal to the forecast window, this was done by subtracting the previous accumulation from the current accumulation. GEFS forecasts have a 3-hourly resolution for the first 96 hours, but a 6-hour accumulation window. As such, every second forecast had to be altered to subtract the forecast before it, yielding 3-hour accumulations. GEFS forecasts were converted to hourly rain rates by dividing the 3-hourly forecasted accumulated precipitation by 3.